%% file: main.tex
\documentclass[journal]{IEEEtran}
\usepackage{graphicx}
\usepackage{float}
\usepackage{amsthm}
\usepackage{amsmath,amsfonts}
\usepackage{amssymb}
\usepackage{caption}
\usepackage{subcaption}
\usepackage{algorithm, algpseudocode}
\usepackage{array}
\usepackage{booktabs}
\usepackage{textcomp}
\usepackage{stfloats}
\usepackage{url}
\usepackage{verbatim}
\usepackage{tikz}
\usepackage{pgfplots}
\usepackage{cite}
\usepackage{color, colortbl}
\usepackage{multirow}
\usepackage{hyperref}
\hypersetup{
    colorlinks = true,    
    linkcolor  = blue,    
    citecolor  = blue,    
    urlcolor   = blue     
}
\pgfplotsset{compat=1.18}

\newcommand{\R}{\mathbb{R}}
\newcommand{\h}{0mm}
\newcommand{\hh}{0mm}

\newcommand{\vs}{0mm}

\input{zoomboxes}

\newcommand\phantomimage{%
    \phantom{%
        \rule{\imagewidth}{\imageheight}%
    }%
}
\newcommand\zoombox[2][]{
    \begin{scope}[zoombox paths]
        \pgfmathsetmacro\xpos{
            (\columncount-1)*(\imagewidth / \pgfkeysvalueof{/tikz/zoomboxarray columns} + \pgfkeysvalueof{/tikz/zoomboxarray inner gap} / \pgfkeysvalueof{/tikz/zoomboxarray columns} ) + \pgflinewidth
        }
        \pgfmathsetmacro\ypos{
            (\rowcount-1)*( \imageheight / \pgfkeysvalueof{/tikz/zoomboxarray rows} + \pgfkeysvalueof{/tikz/zoomboxarray inner gap} / \pgfkeysvalueof{/tikz/zoomboxarray rows} ) + 0.5*\pgflinewidth
        }
        \edef\dospy{\noexpand\spy [
            #1,
            zoombox paths/.append style={
                black and white pattern=\patternnumber
            },
            every spy on node/.append style={#1},
            x=\imagewidth,
            y=\imageheight
        ] on (#2) in node [anchor=north west] at ($(zoomboxes container.north west)+(\xpos pt,-\ypos pt)$);}
        \dospy
        \pgfmathtruncatemacro\pgfmathresult{ifthenelse(\columncount==\pgfkeysvalueof{/tikz/zoomboxarray columns},\rowcount+1,\rowcount)}
        \global\let\rowcount=\pgfmathresult
        \pgfmathtruncatemacro\pgfmathresult{ifthenelse(\columncount==\pgfkeysvalueof{/tikz/zoomboxarray columns},1,\columncount+1)}
        \global\let\columncount=\pgfmathresult
        \ifblackandwhitecycle
            \pgfmathtruncatemacro{\newpatternnumber}{\patternnumber+1}
            \global\edef\patternnumber{\newpatternnumber}
        \fi
    \end{scope}
}

\begin{document}
\title{Multispectral Texture Synthesis using RGB Convolutional Neural Networks}

\author{Sélim~Ollivier, Yann~Gousseau, and~Sidonie~Lefebvre
\thanks{S. Ollivier is with the LTCI, Télécom Paris, Institut Polytechnique de Paris, 91120 Palaiseau, France and also with DOTA/AILab, ONERA, Université Paris-Saclay, 91120, Palaiseau, France (e-mail: selim.ollivier@onera.fr).}
\thanks{Y. Gousseau is with the LTCI, Télécom Paris, Institut Polytechnique de Paris, 91120 Palaiseau, France.}
\thanks{S. Lefebvre is with the DOTA/AILab, ONERA, Université Paris-Saclay, 91120, Palaiseau, France.}}

\markboth{}
{Shell \MakeLowercase{\textit{et al.}}: Bare Demo of IEEEtran.cls for Journals}

\maketitle

\begin{abstract}
    State-of-the-art RGB texture synthesis algorithms rely on \textit{style distances} that are computed through statistics of deep features. These deep features are extracted by classification neural networks that have been trained on large datasets of RGB images. Extending such synthesis methods to multispectral images is not straightforward, since the pre-trained networks are designed for and have been trained on RGB images. In this work, we propose two solutions to extend these methods to multispectral imaging. Neither of them require additional training of the neural network from which the second order neural statistics are extracted. The first one consists in optimizing over batches of random triplets of spectral bands throughout training. The second one projects multispectral pixels onto a 3 dimensional space. We further explore the benefit of a color transfer operation upstream of the projection to avoid the potentially abnormal color distributions induced by the projection. Our experiments compare the performances of the various methods through different metrics. We demonstrate that they can be used to perform exemplar-based texture synthesis, achieve good visual quality and comes close to state-of-the art methods on RGB bands.
\end{abstract}


\IEEEpeerreviewmaketitle

\section{Introduction}

\IEEEPARstart{I}n recent years, a great deal of effort has been devoted to the development and design of multispectral imagers, capable of capturing images of a scene in multiple (usually between 2 and 20) spectral bands in the infrared or visible range. These devices serve various purposes, such as detecting aircraft and UAVs, and analyzing satellite data for Earth observation, including vegetation, atmosphere, ocean, and land assessments. Moreover, numerous deep learning methods have been developed in response to growing interest for multispectral imaging and its applications. 

Yet, the training of efficient classifiers and detection systems requires access to a large amount of data. Similarly, the augmentation of existing databases is highly beneficial for the performance assessment of image processing algorithms and optical sensors. In both cases, it is necessary to have reference scenes and to be able to model the variability of objects of interest and background environments. Given the significant costs associated with data acquisition and annotation, the ability to rapidly synthesize images that accurately replicate the radiometric properties and textures of real-world backgrounds is highly advantageous.

Texture modeling and analysis was initially limited to mono-band images for which tools such as local statistics extraction, co-occurrence matrices, wavelet transforms or fractal based techniques have been developed to account for the spatial behaviour and patterns. Extensions of these techniques to color and multi/hyperspectral images followed afterwards. In this context, the color or spectral statistics need to be taken into account. For this purpose, dimensionality reduction \cite{mercier_characterization_2002} or 3D power spectral density and histogram matching \cite{sarkar_hyperspectral_2010} have been introduced for texture characterization or synthesis. 

Since the foundational work of Gatys et al \cite{gatys_texture_2015}, pre-trained convolutional neural networks (CNN) have been the state-of-the-art in the field of RGB exemplar-based texture synthesis. Most subsequent works used statistics extracted from the feature maps of these networks to model texture through {\it style distances}. This approach is particularly effective because it capitalizes on the network’s powerful description and recognition abilities, which stem from training on large datasets.
Recently, generative networks for the synthesis of satellite data and multi/hyperspectral images have emerged. These models are often based on architectures that have been proven to be effective for RGB images and manage to generate realistic multispectral images by training new neural networks on multi/hyperspectral images \cite{li_generating_2020}\cite{mohandoss_generating_2020}\cite{liu_diverse_2023}\cite{alibani_multispectral_2024}. However, to the best of our knowledge, no method has been proposed to perform multi/hyperspectral texture synthesis that leverage style distances relying on statistics extracted from pre-trained networks as those used for RGB images. Indeed, This is not straightforward, as the widely used VGG19 network, trained on ImageNet, is designed for 3-channel images. Although interesting works have tried to bypass pre-trained networks as VGG-19 to define style distances, results remain of lower quality \cite{ustyuzhaninov_texture_2016}. Moreover, training or fine-tuning a new network would need large multi/hyperspectral texture databases. 

To avoid this, the present work introduces two different strategies to build multispectral style distances based on pre-trained CNN. These distances are then successfully used to perform multispectral texture synthesis. 
Our approach can be applied to a wide range of natural textures. Our numerical analyses target cloud backgrounds, that are useful for various applications : cloud cover forecasting to inform acquisition decisions, and thus optimize the planning of agile satellite constellations for Earth observation, but also to predict the availability of optical communication links between ground stations and satellites. Additional use cases involve the detection of small objects, such as UAVs, against a clouded sky background. Another important application area lies in the context of cloud removal in optical satellite images, which requires pairs of images with and without clouds for training algorithms. Czerkawski et al. \cite{czerkawski_satellitecloudgenerator_2023} designed an open-source python toolbox based on Perlin noise enabling the synthesis of diverse simulated pair data, with controllable parameters to adjust cloud appearance. This tool has been evaluated as a source of training images supply for cloud detection and cloud removal algorithms, and has led to good performances compared to the same algorithms trained only on real data.\\

Our contributions can be summarized as the following:

\begin{enumerate}
    \item we introduce two style distances: a stochastic style distance and a projected style distance that can be computed from a a pre-trained RGB CNN. This in turn allows to synthesize multispectral images using an optimization process similar to the one of \cite{gatys_texture_2015}. Our approach is not specific to texture modeling and could also be used to extend other distances based on pre-trained RGB networks to multispectral imaging without having to train new networks for that purpose.
    \item to correct the color domain shift induced by the projection, we propose to rely on color transfer, meant to align the projected image with a more typical color distribution seen during training by the pre-trained CNN involved in the style distance. We then empirically investigate its usefulness.
\end{enumerate}
\vspace{.3cm}
Section II reviews related work on multispectral texture modeling, and section III presents our proposed methodology in detail. Texture synthesis experiments on 130 multispectral cloud field images from the Sentinel-2 mission are conducted in section IV. We conclude this paper in section V.

\section{Related work}
\paragraph{Multi and hyperspectral texture modeling}

Texture synthesis and modeling first emerged in applications linked to monochromatic images and extensions to colour images were proposed later on. For color texture synthesis, considering each color channel independently is often not satisfying due to inter-channel correlations. Early methods for color texture synthesis apply a Principal Component Analysis (PCA) before working channel-by-channel~\cite{heeger_pyramid-based_1995}, possibly modeling inter-channel correlations as in~\cite{portilla_parametric_2000}. For hyperspectral images, Mercier et al. \cite{mercier_characterization_2002} uses an Independent Component Analysis before applying wavelet decomposition for texture characterization. Sarkar et al. \cite{sarkar_hyperspectral_2010} proposes to constrain the multi-dimensional histograms of images together with a 3D Fourier transform accounting for intra-band spatial dependency to perform texture synthesis. 

From a modeling perspective, textures were originally often analysed through dependencies between nearby pixels, which led to the use of Markov models. 
Adding the assumption of a Gaussian response for each vertex of the graph representing the image to be analysed led to a model well-known as a random Gauss-Markov field. These models were first leveraged for anomaly detection in multispectral and hyperspectral images by Schweizer and Moura \cite{schweizer_hyperspectral_2000}\cite{schweizer_efficient_2001}: the inter-pixel Markov component allows textures to be taken into account while the underlying Gaussian assumption allows a simple modeling of the spectral response. Some Gauss-Markov models with different way of accounting for second order spectral information were proposed by Rellier et al \cite{rellier_gauss-markov_2002}\cite{rellier_texture_2004}, for hyperspectral textures classification. All these models are based on sets of simplifying assumptions taking into account the properties of the hyperspectral images analysed. These assumptions relate in particular to the constancy across the spectrum of the Markov parameters governing the textures and on the spectral covariance modeling of the textures across the spectrum. Spatial parameters are typically assumed to be constant throughout the spectrum, which means that the multispectral texture is considered to be identical at each wavelength, up to a factor of variance. The Schweizer and Moura formulation implies that each scalar pixel interacts with its neighbours in all three dimensions (spatial and spectral) and the inclusion of an inter-band Markov link in the model implicitly imposes identical inter-band behaviour between all pairs of neighbouring bands. Rellier et al proposed to restrict the pixel neighbourhood to spatial neighbours. The aim is then to model the spatial interactions between hyperpixels, and thus the expression of the spatial part of the precision matrix loses its interband component.
Although more flexible variations have been proposed, these underlying simplifying assumptions limit the expressiveness of the textures generated, and texture synthesis methods based on generative networks offer a promising alternative. 

To assess the quality of the synthesis, conventional texture metrics, such as local binary pattern and co-occurrence matrices, have been exploited, with within and between channel interactions, in a recent comparison of color imaging and hyperspectral imaging for texture classification \cite{porebski_Comparison_2022}.
Recently, Chu and co-authors focused on accounting simultaneously for spectral and spatial distribution. They proposed a four dimensional texture metrics, GHOST \cite{chu_ghost_2021}, encompassing joint metrological assessment of spectral and spatial properties. They also conceived an alternative feature, Relative Spectral Difference Occurrence Matrix (RSDOM) \cite{chu_metrological_2019}\cite{chu_hyperspectral_2021}, which considers the joint probability of spectral distribution and of pixels spatial arrangement and consists of pixel differences with spectral reference and spatial neighbors, and emphasized its contribution in the hyperspectral image texture classification scheme on HyTexiLa database \cite{khan_hytexila_2018}.\\

\paragraph{Deep generative models for multispectral and hyperspectral images}
The most recent generative neural networks methods such as GANs (Generative adversarial Networks) and diffusion models have only very recently started to be applied on multispectral satellite data. The recent work of \cite{alibani_multispectral_2024} compares, both with a spectral signature analysis and a modified Fréchet inception distance, the performances of StyleGAN3 and proGAN architectures to reproduce Sentinel-2 images. Khanna et al. \cite{khanna_diffusionsat_2024} propose DiffusionSat, a generative foundation model for remote sensing data based on the latent-diffusion model architecture of StableDiffusion \cite{rombach_high-resolution_nodate}, dedicated to inverse problems such as multispectral input super-resolution, temporal prediction, and in-painting. In order to alleviate the computational complexity due to the high spectral dimensionality, UnmmixDiff \cite{yu_unmixdiff_2024} has recently been proposed, and consists in coupling an unmixing autoencoder and a diffusion model. The synthesis if thus performed in the abundance space, which enables to obtain realistic objects' distribution in abundance maps. However, this method focuses on the spectral profiles consistency rather than on the texture representativity of the different natural background components, such as forest, sky, grassland...  
Another publication \cite{liu_diverse_2023} emphasizes the use of a diffusion model for the synthesis of hyperspectral images, but with a conditional RGB image as input. They introduce a conditional vector quantized generative adversarial network (VQGAN) to map hyperspectral data into a low-dimensional latent space, in which the diffusion model is easier to train. They pay particular attention to the diversity of the generated hyperspectral images, by proposing two new metrics focusing on spectral content diversity. The syntheses are evaluated using both pixel-wise metrics such as RMSE, spectral angle mapper (SAM), and PSNR, and the structure similarity index metric SSIM for evaluating the quality of spatial information. 
Additionally, foundation models—trained on vast unlabelled datasets and fine-tuned for specific tasks using limited annotated data—have seen increased use in satellite imagery over the past two years. FG-MAE \cite{wang_feature_2023} and SatMAE \cite{cong_satmae_2022} have been developed for multispectral Sentinel-2 images, and DOFA \cite{xiong_neural_2024} has been very recently proposed to adaptively integrate various data modalities into a single framework, leveraging the concept of neural plasticity in brain science. These foundation models will undoubtedly contribute to strengthen multispectral image analysis capabilities.\\
These models have only been applied on satellite data, and mostly on data taken over cities, not on textured natural backgrounds. \\

\paragraph{Neural texture synthesis} 

The work from Gatys et al.~\cite{gatys_texture_2015} was a major breakthrough for exemplar-based texture synthesis. The main idea of this work, that we will detail in Section \ref{subsec:gatys}, is to constrain some statistics on deep image features extracted by classical classification networks that have been trained on large image datasets (such as the VGG19 network trained on ImageNet). More precisely, second order statistics between deep features are computed on an exemplar image through Gram matrices and images are then synthesized by optimizing the pixel responses to match these matrices. This results in very realistic synthesized texture images, although at the price of a relatively heavy optimization procedure for each synthesis. Many works have followed, all exploiting the central idea of measuring distance between texture images through the squared difference between their Gram matrices, distances that in this paper we will call {\it style distances}. A particular interest was devoted to develop feed-forward architecture to speed-up the synthesis~\cite{ulyanov_texture_2016}, use the style distances in the GAN framework \cite{jetchev_texture_2017} or to develop auto-encoder texture synthesizers~\cite{chatillon_geometrically_2023}. To the best of our knowledge, none of these state-of-the-art texture synthesis approaches have been adapted to the multispectral or hyperspectral framework, which is the purpose of the present paper. 

\section{Methods}

\subsection{Neural statistics for texture synthesis}
\label{subsec:gatys}

As recalled in the previous section, the method from Gatys et al.~\cite{gatys_texture_2015} rely on the use of statistics of deep features. These features are classically obtained from the VGG19 network. It consists of a succession of convolutional layers. Each layer $l$ is made of $N_l$ convolutional filters and provides a feature map $F^l \in \mathbb{R}^{N_l\times M_l}$ (where $M_l$ is the number of pixels of the feature map). The Gram matrices were chosen by the authors to form a summary statistic that is blind to spatial information:

\begin{equation}
    \forall l \in \{1,...,L\},~~~~G^l = F^l \left(F^l\right)^T.
\end{equation}

The set $G = \left\{G^1,...,G^L\right\}$ of Gram matrices computed from the network in response to an image constitutes a stationary multi-scale description of the texture. It can further be used to define a style distance between two images $I$ and $\hat{I}$ with Gram matrices $G$ and $\hat{G}$:

\begin{equation}\label{eq:gatys}
    \mathcal{L}_{style}\left(I,\hat{I}\right) = \sum_l w_l \left(G^l - \hat{G}^l\right)^2,
\end{equation}

where the $w_l$ are chosen weighting factors.\\

This style distance was used in~\cite{gatys_texture_2015} to achieve a clear breakthrough for texture synthesis. It was also later used to train other neural networks as a loss function \cite{ulyanov_texture_2016}\cite{lin_towards_2022}\cite{chatillon_geometrically_2023}. In this work, we follow the exemplar synthesis process presented by \cite{gatys_texture_2015}:  given an exemplar image $I$, a synthetic image $\hat{I}$ is initialized with a Gaussian white noise and then gradient descent is used to minimize the style distance between the two images $I$ and $\hat{I}$. The result of this optimization is expected to match the Gram matrices of the exemplar. In the following, we chose to initialize the synthetic image following a Gaussian noise with the mean and covariance of the original image as it showed better performances.

This style distance makes use of the Gram matrices to describe the feature maps distributions, but other statistics as covariance matrices (i.e. Gram matrices of the centered feature maps) have also shown good performances for texture synthesis. 

We chose to use the latter in our work.

The main obstacle to the extension of CNN based texture synthesis processes to multispectral imaging is the design and the training of a multispectral CNN from which to extract feature maps statistics. To address this, we introduce a stochastic style distance in section \ref{subsec:stochastic_style_distance} and a style distance that makes use of projection in section \ref{subsec:projected_style_distance}. In section \ref{subsec:color_transfer}, we add a color operation to the projection to align the color distributions with those of the images the network has been trained on. These distances enables one to use the VGG19 network, pre-trained on ImageNet, that has become standard for 3-channels images.

\subsection{Stochastic style distance}\label{subsec:stochastic_style_distance}

The difficulty to extend the approach recalled in the previous paragraph to multispectral images is to constrain the statistics of an image of dimension $N$ with an optimization procedure that relies on a RGB CNN. To do so, we decide to constrain the statistics of all possible 3-channels images obtained from the channels of the multispectral image. 

We consider an exemplar multispectral image $I$ with $N$ spectral bands and $\hat{I}$ its synthetic counterpart. We propose to randomly draw a triplet $J$ of integers in $\{1,...,N\}$ at each iteration of the optimization. Then, to evaluate the style distance $\mathcal{L}_{style}\left(I_J,\hat{I}_J\right)$ between the images $I_J$ and $\hat{I}_J$ made of the spectral bands indexed by the integers of $J$. Finally, the gradient of this quantity is used to take an optimization step on $\hat{I}$.

During the optimization, one can average the style distance over batches of uniformly drawn triplets in a way that is reminiscent with stochastic gradient descent algorithms. It ensures to take into account more spectral bands and correlations at each iterations. Moreover, it regularizes the objective function.

The global minimization objective of our algorithm is the expectation of the style distance between 3-channels images corresponding to the spectral bands indexed by a triplet $J$ that follows the uniform law. This amounts to averaging the style distance over all possible triplets :

\begin{align}
    \mathcal{L}_{style}^{MS}\left(I, \hat{I}\right)
    &=
    \mathbb{E}_{J \sim \mathcal{U}(\mathcal{J}_N)} 
    \left[
    \mathcal{L}_{style} \left(I_J, \hat{I}_J\right)
    \right]\\
    &=
    \frac{1}{N_J}
    \sum_{J\in\mathcal{J}_N}
    \mathcal{L}_{style} \left(I_J, \hat{I}_J\right)
\end{align}
\label{eq:stochastic_style_distance}

Where :
\begin{itemize}
    \item $\mathcal{J}_N$ is the set of triplets of integers in $\{1,...,N\}$ of cardinal $N_J$,
    \item $\mathcal{U}(\mathcal{J}_N)$ is the uniform distribution on $\mathcal{J}_N$,
    \item $\mathbb{E}_{J \sim \mathcal{U}(\mathcal{J}_N)}$ is the mathematical expectation under the condition that $J$ follows a uniform distribution, 
    \item if $J$ is a triplet, $I_J$ is the 3-channels image made of the 3 spectral bands of $I$ indexed by $J$,
    \item $\mathcal{L}_{style}$ is the classical style loss for RGB images using a VGG19 pre-trained network described by \cite{gatys_texture_2015}, recalled in Equation (\ref{eq:gatys}).
\end{itemize}
\vspace{.3cm}

The proposed synthesis algorithm is detailed in Algorithm \ref{alg:sto_syle} for $n$ iterations of the gradient descent algorithm and $B$ triplets drawn at every iteration. The mean and covariance of $I$ are denoted as $\mu_I$ and $\Sigma_I$.

\begin{algorithm}
    \caption{Texture synthesis with stochastic style loss}
    \label{alg:sto_syle}
    \begin{algorithmic}
        \State $\hat{I} \sim \mathcal{N}\left(\mu_I, \Sigma_I\right)$ \Comment{Gaussian noise initialization}
        \For{$i \in \{1,...,n\}$}
        \State $\mathcal{L} \gets 0$
        \For{$j \in \{1,...,B\}$}
        \State $J \sim \mathcal{U}(\mathcal{J}_N)$ \Comment{Uniformly drawn triplet}
        \State $\mathcal{L} \gets \mathcal{L} + \mathcal{L}_{style}\left(I_J, \hat{I}_J\right)$
        \EndFor
        \State Take a gradient step on $\hat{I}$ with gradient $\nabla \mathcal{L}$
        \EndFor
        \State \Return $\hat{I}$
    \end{algorithmic}
\end{algorithm}

Our method ensures to browse every spectral bands and consider the correlations between each of them in the course of the optimization. However, a drawback of this method is that it may need a relatively large number of triplets to reach a good overall control over multi-spectral statistics, resulting in a high computational cost. In the next section, we introduce a more simple and faster alternative approach. 

\subsection{Projected style distance}\label{subsec:projected_style_distance}

The second style distance that we propose aims to take advantage of the redundancy of the spectral information. Indeed, the spectral information that is contained in $N$ dimensions often can be reduced in a lower number of dimensions.
Therefore, we consider a projector $P : \R^N \to \R^3$ that maps multispectral images with $N$ spectral bands into RGB-like images with 3-channels. The proposed style distance consists then in the classical style distance computed on projected images :

\begin{equation}
    \mathcal{L}_{style}^{P}\left(I, \hat{I}\right)
    =
    \mathcal{L}_{style}\left(P(I), P(\hat{I})\right)
\end{equation}\label{eq:proj_style_dist}

The use of the projected style distance for exemplar-based texture synthesis is detailed in algorithm \ref{alg:proj_syle}.
\begin{algorithm}
    \caption{Texture synthesis with projected style loss}
    \label{alg:proj_syle}
    \begin{algorithmic}
        \State $\hat{I} \sim \mathcal{N}\left(\mu_I, \Sigma_I\right)$ \Comment{Gaussian noise initialization}
        \For{$i \in \{1,...,n\}$}
        \State $I_P \gets P(I)$
        \State $\hat{I}_P \gets P(\hat{I})$
        \State $\mathcal{L} \gets \mathcal{L}_{style}\left(I_P, \hat{I}_P\right)$ 
        \State Take a gradient step on $\hat{I}$ with gradient $\nabla \mathcal{L}$
        \EndFor
        \State \Return $\hat{I}$
    \end{algorithmic}
\end{algorithm}

\subsection{Color transfer for neural statistics extraction}\label{subsec:color_transfer}

We observe that the use of a projection can lead to unusual color distributions that can have an influence on the quality of the description provided by the feature maps as the network may not have been trained on such distributions. This domain shift can affect the quality of the description provided by the CNN. Examples of the color that are induced by a PCA projection for cloud field images taken from Sentinel-2 mission are available in Figure \ref{fig:pca_ex}.

To circumvent this, we propose to perform a color transfer operation after the projection and before the use of the VGG-19 network. The target distribution of this transfer operation is simply the color distribution of a natural image, that therefore acts as a color palette. The idea here is to  send back the considered 3-channels image in regions of the color space that are likely to be closer to the regions that have been explored during the training of the network. 

\renewcommand{\h}{0.32\linewidth}
\begin{figure}
    \centering
    \includegraphics[width=\h]{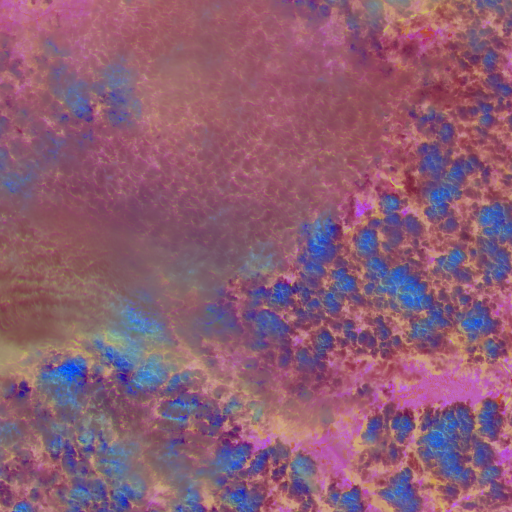}
    \includegraphics[width=\h]{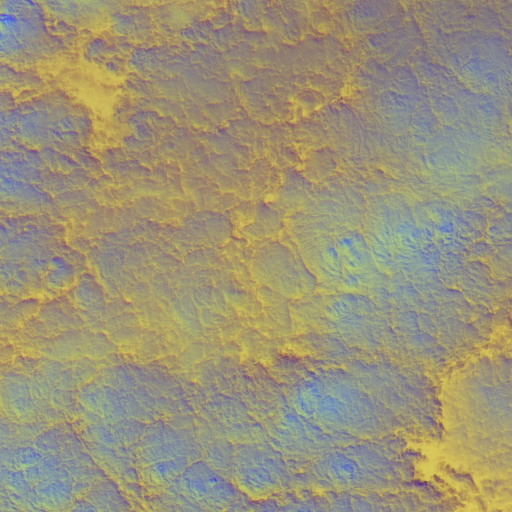}
    \includegraphics[width=\h]{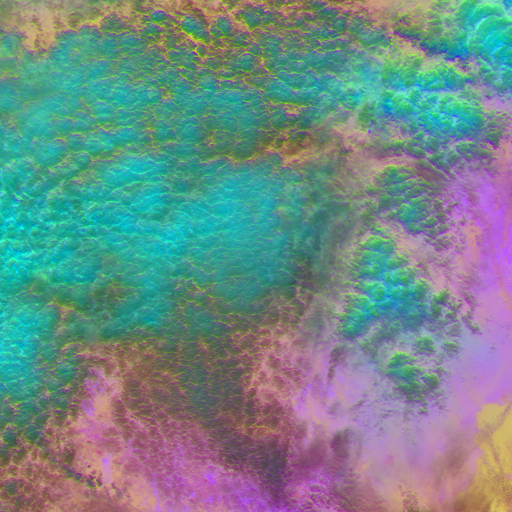}
    \caption{Examples of unusual color distributions induced by the PCA projection of Sentinel-2 cloud field images.}
    \label{fig:pca_ex}
\end{figure}

\begin{figure}
    \centering
    \resizebox{\linewidth}{!}{\begin{tikzpicture}
        \node (C) at (3.5,7) {\includegraphics[width=2cm]{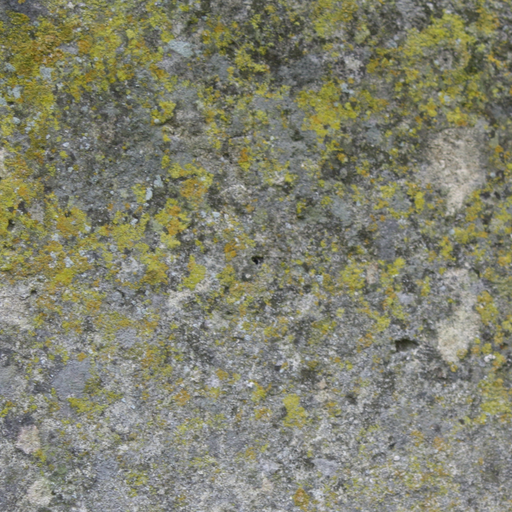}};
        
        \draw[fill=gray!30] (-2,4) rectangle (0,6);
        \draw[fill=gray!30] (-1.8,3.8) rectangle (0.2,5.8);
        \draw[fill=gray!30] (-1.6,3.6) rectangle (0.4,5.6);
        \draw[fill=gray!30] (-1.4,3.4) rectangle (0.6,5.4);
        \draw[fill=gray!30] (-1.2,3.2) rectangle (0.8,5.2);
        \node (A) at (0,4) {\includegraphics[width=2cm]{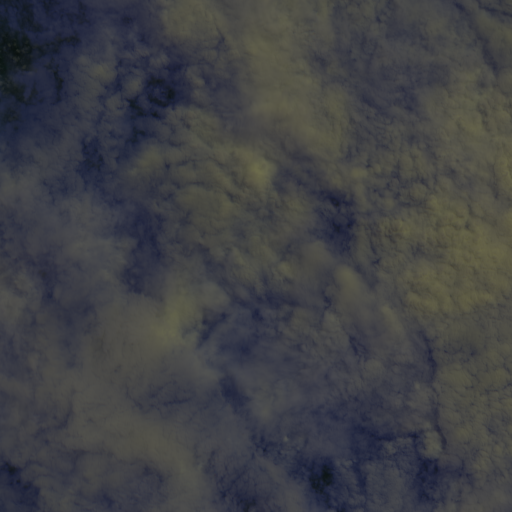}};
        \node (AP) at (3.5,4) {\includegraphics[width=2cm]{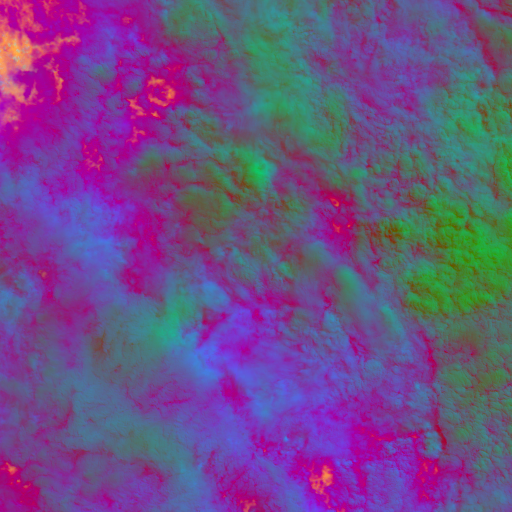}};
        \node (AC) at (7,4) {\includegraphics[width=2cm]{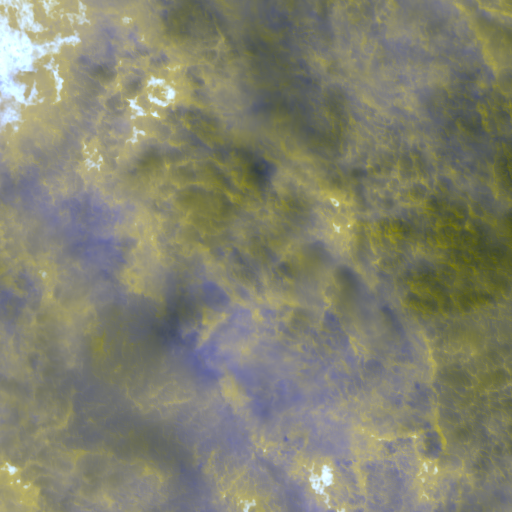}};

        \draw[fill=gray!30] (-2,0) rectangle (0,2);
        \draw[fill=gray!30] (-1.8,-0.2) rectangle (0.2,1.8);
        \draw[fill=gray!30] (-1.6,-0.4) rectangle (0.4,1.6);
        \draw[fill=gray!30] (-1.4,-0.6) rectangle (0.6,1.4);
        \draw[fill=gray!30] (-1.2,-0.8) rectangle (0.8,1.2);
        \node (B) at (0,0) {\includegraphics[width=2cm]{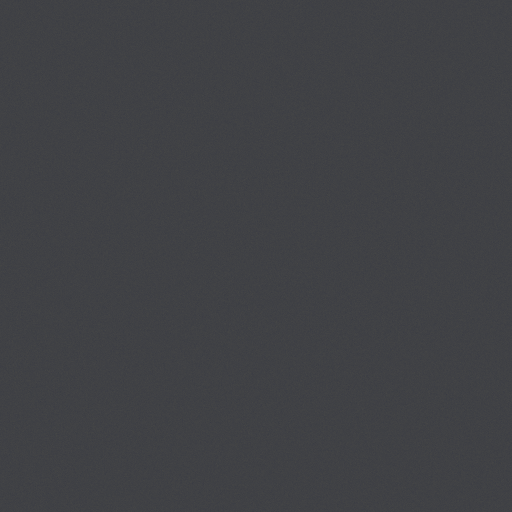}};
        \node (BP) at (3.5,0) {\includegraphics[width=2cm]{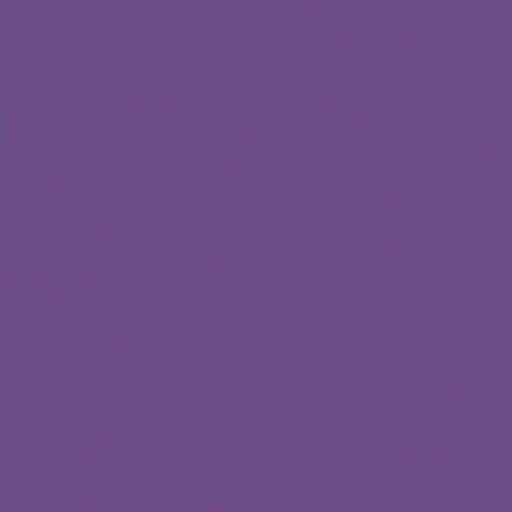}};
        \node (BC) at (7,0) {\includegraphics[width=2cm]{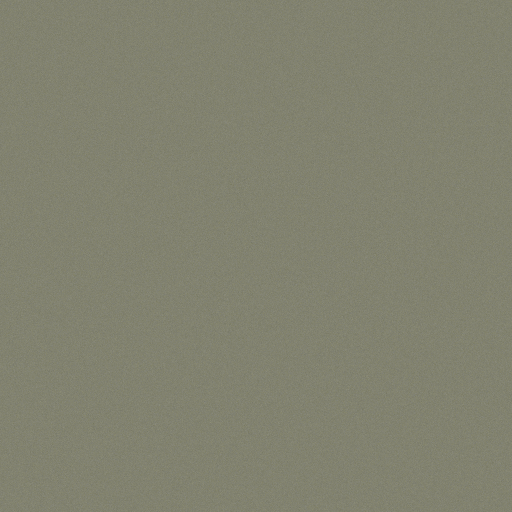}};

        \draw[fill=orange!30] (9,3.15) -- (10.5,2.65) -- (10.5,1.35) -- (9,0.85) -- cycle;
        \node at (9.75,2) {VGG19};

        \node (G) at (11.5,2.4) {$G$};
        \node (GH) at (11.5,1.6) {$\hat{G}$};

        \draw[thick,->] (A) -- (AP) node[midway, below] {$P$}; 
        \draw[thick,->] (AP) -- (AC) node[midway, below] {$T_C$}; 
        \draw[thick,->] (B) -- (BP) node[midway, below] {$P$};
        \draw[thick,->] (BP) -- (BC) node[midway, below] {$T_C$}; 
        \draw[thick,-] (C) -| (5.25,4);
        \draw[thick,-] (AC) -| (8.5,2.4);
        \draw[thick,-] (BC) -| (8.5,1.6);
        \draw[thick,->] (8.5,2.4) -- (8.9,2.4);
        \draw[thick,->] (8.5,1.6) -- (8.9,1.6); 
        \draw[thick,->] (10.5,2.4) -- (G);
        \draw[thick,->] (10.5,1.6) -- (GH); 
    \end{tikzpicture}}
    \caption{Scheme of the proposed use of color transfer for multispectral texture synthesis. A projection $P$ is used to obtain 3-channels images from the exemplar and synthetic multispectral images. Then, a color transfer operator $T_C$ is applied to both images to match their color statistics to the one of a reference palette image. 
    Finally, statistics of deep features are extracted from the so-obtained images to perform texture synthesis.}
    \label{fig:gatys_pca_color}
\end{figure}
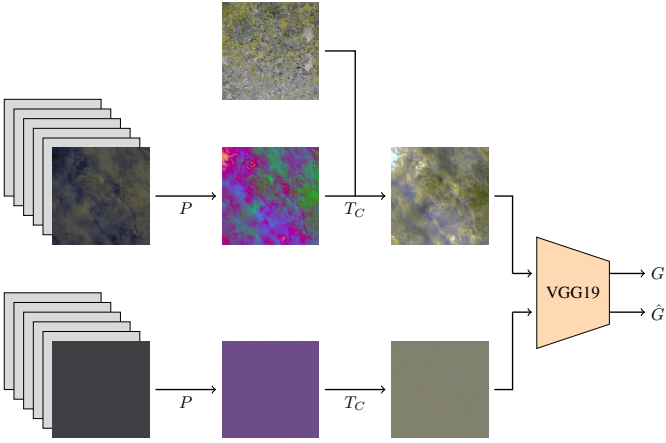

To define such a color transfer operation, we chose to describe the color distributions of the images by the first and second order statistics, assuming that it gives a first satisfactory and sufficient approximation for our application. Hence, we consider an affine transformation that maps the projected image $I_P$ with color statistics $(\mu_P, \Sigma_P)$ to an image matching the color statistics $(\mu_C, \Sigma_C)$ of a palette image $I_C$. We denote $T_{P\to C}$ such a transformation and, for an image $I$, we have: 

\begin{equation}\label{eq:color_transformation}
    T_{P\to C}(I) 
    = 
    \mu_C + L_C L_P^{-1} \left(I - \mu_P\right)
\end{equation}

where the $L_P$ and $L_C$ matrices are obtained using Cholesky decomposition of covariances matrices:
\begin{align}
    \Sigma_P &= L_P L_P^T\\
    \Sigma_C &= L_C L_C^T
\end{align}

As it is possible to define the converse transformation $T_{C\to P}$, such a transformation happens to be reversible so that it conserves the spatial information of the images and only act on the color of the pixels to match the first and second order statistics.

\renewcommand{\h}{0.85\linewidth}
\begin{figure*}
    \centering
    \includegraphics[width=\h]{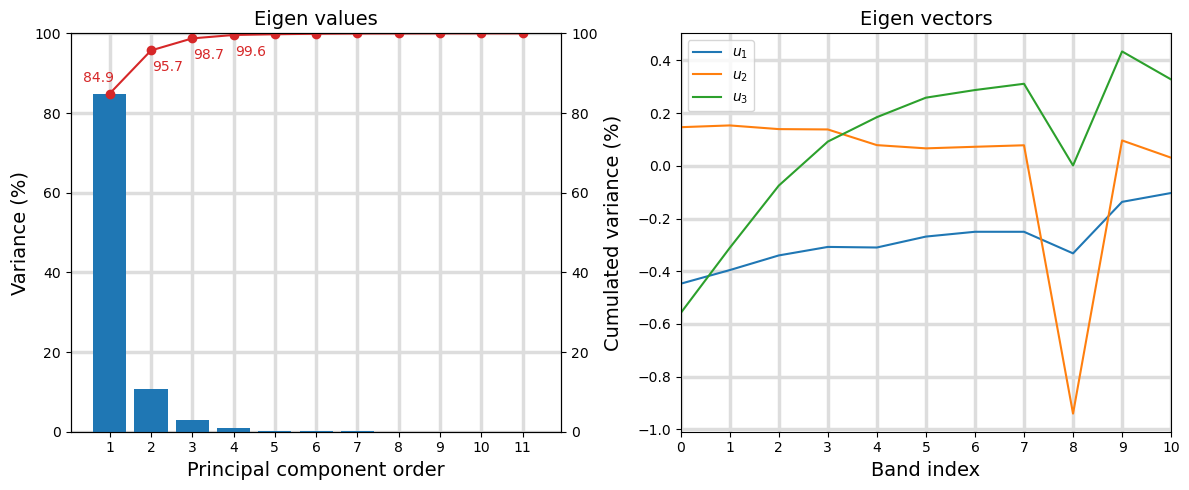}
    \caption{(Left) Eigenvalues of the used PCA over 11-band Sentinel-2 images. The first component encompasses $95\%$ of the variance, the 3 first $99\%$. (Right) 3 first eigenvectors of the used PCA over 11-band Sentinel-2 images (denoted as $u_i$).}
    \label{fig:pca}
\end{figure*}

The corresponding style distance is therefore (see Figure \ref{fig:gatys_pca_color} for illustration):  
\begin{align}
    \mathcal{L}_{style}^{P+C}\left(I, \hat{I}\right)
    &=
    \mathcal{L}_{style}\left(T_{P\to C}(P(I)), T_{P\to C}(P(\hat{I}))\right)\\
    &=
    \mathcal{L}_{style}\left(P_C(I), P_C(\hat{I})\right)
\end{align}\label{eq:proj_Color_style_dist}

with $T_{P\to C}$ being the color transformation introduced earlier, that maps the color statistics of the projection of the exemplar texture $P(I) = I_P$ into the color statistics of the palette color image $I_C$, and $P_C = T_C \circ P$ being the new projection operator.

\subsection{Multispectral texture synthesis with deep features statistics}
To perform multispectral texture synthesis, we initialize a synthetic multispectral image of $N$ channels with a Gaussian noise of mean and covariance of the exemplar image as it showed better results than a white noise initialization. Then, we use L-BFGS \cite{liu_limited_1989} to update the synthetic multispectral image in order to minimize a style distance. At each iteration of the algorithm, the gradients of one of the aforementioned style distances are used to update the pixels of the $N$ channels synthetic image. The style distance uses a CNN trained on RGB images to extract deep features statistics, covariance matrices here.

\section{Experiments}
In this section, we give a brief description of the dataset we consider, the experimental settings and the metrics used to compare the texture syntheses. Comprehensive comparisons of syntheses using the different style distances that we formerly introduced as objective functions are then presented and discussed. 

\subsection{Dataset}\label{sec:data}

We consider a dataset composed of 130 512$\times$512 Sentinel-2 images of cloud fields. The Sentinel-2 mission is an Earth observation mission from the Copernicus Program that acquires optical imagery at a spatial resolution ranging from 10 m to 60 m over land and coastal waters. It captures images in 12 spectral channels in the visible/near infrared (VNIR) and short wave infrared spectral range (SWIR), including 9 proper observation bands and 3 atmospheric correction bands (near 442, 945 and 1375 nm), for which the atmospheric absorption is high. The last band at 1375 nm present particularly high absorption percentage and is not exploitable. As a consequence, only 11 spectral bands are available.

\subsection{Experimental settings} 

We use the different style distances that we introduced in this paper as objective functions to perform texture synthesis on the 11-band images. Additional results over the 9 observation bands are also available in Appendix \ref{appendix:9bands}. 

Experimental choices and hyperparameters for each objective are as follows:
\begin{itemize}
    \item \textbf{Stochastic} : we use a batch size of 10 triplets at each iterations. We show further experiments varying the batch size in Appendix \ref{appendix:sto_batch_size}.
    \item \textbf{PCA} : we use Principal Component Analysis to encode the spectral information contained in 11 bands into 3 dimensions. In our case, these 3 first components are sufficient to explain $99\%$ of the variance (see Figure \ref{fig:pca} for eigenvalues and eigenvectors).
    \item \textbf{PCA + Color} : We use 3 different color exemplar images to perform color transfer upstream of the PCA projection : \textit{Wall}, \textit{Pebbles} and \textit{Fabrics}. The images are available in Figure \ref{fig:ex_img_ms}. 
\end{itemize}

\renewcommand{\v}{6.5cm}

\renewcommand{\h}{0.32\linewidth}
\begin{figure}
    \centering
    \begin{subfigure}{\h}
        \includegraphics[width=\textwidth]{images/wall_1.png}
        \caption{Wall}
    \end{subfigure}
    \begin{subfigure}{\h}
        \includegraphics[width=\textwidth]{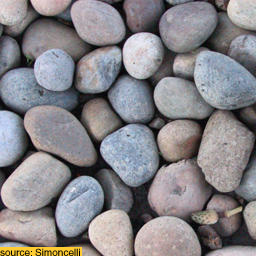}
        \caption{Pebbles}
    \end{subfigure}
    \begin{subfigure}{\h}
        \includegraphics[width=\textwidth]{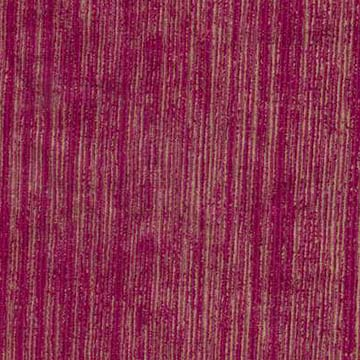}
        \caption{Fabrics}
    \end{subfigure}
    \caption{Exemples of palette images used for the proposed color transfer.}
    \label{fig:ex_img_ms}
\end{figure}

For style distances, we use a VGG19 network initialized with the pre-trained weights from \cite{gatys_texture_2015}. As the authors of the latter, we extract the feature maps of the first convolutionnal filters at each spatial resolution (i.e. Conv layers with indices 1,3,5,8,11) and use quadratic decaying loss weights $w_l = \frac{1}{N_l^2}$ where $N_l$ is the number of feature maps of the layer $l$. 
As mentioned in section \ref{subsec:gatys}, we use the covariance matrices as deep features statistics instead of the Gram matrices and we initialize the synthetic image with a Gaussian noise with the mean and covariance of the exemplar image.
We use L-BFGS for 500 iterations to perform optimization.

\subsection{Metrics}\label{subsec:metrics}

Unlike traditional pixel-by-pixel reconstruction, we want to  generate a new texture image that conserves the statistics, the properties and the general visual aspect of the original image while allowing another spatial distribution. Therefore, traditional pixel-wise metrics such as the PSNR are not adapted. In the following, we introduce different metrics used to evaluate the performance of our methods.\\ 

\subsubsection{Style}
We measure different style distances between original and synthetic images :
\begin{itemize}
    \item the stochastic style distance denoted as $\mathcal{L}_{style}^{MS}$. It is computed by  averaging over every possible triplets. This is precisely the metric corresponding to the stochastic style distance.  
    \item the projected style distance using a PCA, denoted as $\mathcal{L}_{style}^{PCA}$, corresponding to the second style distance introduced in this paper. 
    \item the style distance for the bands 2,3 and 4 which are the closest to RGB wavelengths. It is denoted as $\mathcal{L}_{style}^{RGB}$ 
\end{itemize}
\vspace{.2cm}

\subsubsection{Multispectral distribution}
In the context of RGB images, color distribution is a central feature in texture description. For multispectral images, we similarly want to synthesize images having a distribution of pixel values as close as possible to the exemplar image. An efficient way to compare multidimensional distribution is the Wasserstein distance. Because it is very costly to compute such distances as soon as we are in dimension greater than two, we either rely on the sliced Wasserstein distance \cite{rabin_wasserstein_2012} or to band-wise distances:  
\begin{itemize}
    \item sliced Wasserstein distance $\mathcal{L}_{hist}$:
    \begin{align*}
        \mathcal{L}_{hist}(I,\hat{I})^2 
        &= 
        SW_2(I,\hat{I})^2\\
        &= 
        \mathbb{E}_{v\sim\mathcal{U}\left(\mathbb{S}^{N-1}\right)}
        \left[
        W_2(I \cdot v, \hat{I} \cdot v)^2
        \right]\\
        &=
        \mathbb{E}_{v\sim\mathcal{U}\left(\mathbb{S}^{N-1}\right)}
        \left[\left\|
        \mathrm{sort}(I \cdot v) - \mathrm{sort}(\hat{I} \cdot v)
        \right\|^2_2\right],
    \end{align*}
    where $\mathbb{S}^{N-1}$ is the unit sphere in dimension $N$, $SW_2$ the $2$-sliced Wasserstein distance, $W_2$ the $2$-Wasserstein distance, $I\cdot v$ the scalar product between $I$ and $v$, and $\mathrm{sort}$ the sorting operator.
    \item band-wise Wasserstein distance $\mathcal{L}_{hist}^\lambda$ for a band $\lambda$:
    \begin{align*}
        \mathcal{L}_{hist}^\lambda(I,\hat{I}) 
        &= 
        W_2(I_\lambda,\hat{I}_\lambda)\\
        &= 
        \left\|
        \mathrm{sort}(I_\lambda) - \mathrm{sort}(\hat{I}_\lambda)
        \right\|_2
    \end{align*}
\end{itemize}
\vspace{.2cm}
\subsubsection{Gaussian hypothesis}
Anomaly detection is one of the most common analyses performed on multispectral images. The historical multispectral anomaly detection criterion, the RX \cite{reed_adaptive_1990}, makes a Gaussian assumption for the whole image, or at least for limited windows, in order to define a statistical test. Hence, we also measure the difference in the first and second order spectral statistics, using :
\begin{itemize}
    \item the $L_2$ norm between the means and covariances, respectively $\mathcal{L}_\mu$ and $\mathcal{L}_\Sigma$ :
    \begin{align*}
        \mathcal{L}_\mu(I,\hat{I}) &= \left\|\mu - \hat{\mu}\right\|_2\\
        \mathcal{L}_\Sigma(I,\hat{I}) &= \left\|\Sigma - \hat{\Sigma}\right\|_2
    \end{align*}
    \item the Wasserstein distance between the two gaussian distributions given by the means and covariances which, for the sake of simplicity, we label $RX$ even though it is not equivalent to the $RX$ criterion (we do not compute a Mahalanobis distance):
    \begin{equation}
        \mathcal{L}_{RX}(I,\hat{I}) = \sqrt{(\mu - \hat{\mu})^2 + \mathrm{Tr}\left(\Sigma + \hat{\Sigma} - 2\Sigma^\frac{1}{2}\hat{\Sigma}\Sigma^\frac{1}{2}\right)}
    \end{equation}
    where $I$ and $\hat{I}$ have respectively $(\mu, \Sigma)$ and $(\hat{\mu}, \hat{\Sigma})$ as statistics.
\end{itemize}
\vspace{.2cm}

\subsubsection{Spectrum}
Cloud fields images exhibit fractal behaviors, which can be analyzed through the decay of the azimuthal modulus of the Fourier's transform of the image. which is defined as follows : 
\begin{equation}
    \mathcal{F}_{rad}(I)(r)
    = 
    \frac{1}{N_r} 
    \int_0^{2\pi}
    \left\| 
    \mathcal{F}(I)(r \cos{\theta}, r\sin{\theta})
    \right\|
    d\theta
\end{equation}
where $I$ is a single channel image and $\mathcal{F}$ the Fourier's transform operator.
As one expects a power-law decay $\mathcal{F}_{rad} (r) \sim \frac{1}{r^p}$, the differences in fractal behaviours is quantified by the $l_2$ distance between the two logarithm of the azimuthal spectra :
\begin{equation}
    \mathcal{L}_{sp}(I,\hat{I}) 
    =
    \left\|\log\Big(\mathcal{F}_{rad}(I)\Big) - \log\Big(\mathcal{F}_{rad}(\hat{I})\Big)\right\|_2
\end{equation}
We compute this distance on the band-wise image $I_{\lambda}$ for each spectral band $\lambda$ and on the grey-scale image $I_{mean}$ :
\begin{align}
    \mathcal{L}_{sp}^\lambda(I,\hat{I}) &= \mathcal{L}_{sp}(I_\lambda,\hat{I}_\lambda)\\
    \mathcal{L}_{sp}^{mean}(I,\hat{I}) &= \mathcal{L}_{sp}(I_{mean},\hat{I}_{mean})    
\end{align}
\vspace{.2cm}

\subsubsection{Gradients} The final features we consider are image gradients, which are key to several cloud image applications \cite{szczap_flexible_2014}\cite{moradi_false-alarm_2018}\cite{ge_cloud_2023}. Gradient calculation in the two spatial directions (horizontal and vertical) on an image $I$, leads to two multispectral images $\nabla_x I$ and $\nabla_y I$ that can be used together to obtain the magnitude resulting in another multispectral image $|\nabla I|$. A proper reconstruction should exhibit a distribution of gradients as close as possible to the original one. Hence, we use the sliced Wasserstein distance to evaluate the deviation between the two distributions :
\begin{equation}
    \mathcal{L}_{grad}(I,\hat{I}) 
    = 
    SW_2\left(|\nabla I|, |\nabla\hat{I}|\right)
\end{equation}

We further compute the Wasserstein distance over the spectral bands in a similar way :
\begin{equation}
    \mathcal{L}_{grad}^\lambda(I,\hat{I}) 
    = 
    W_2(\nabla I_\lambda, \nabla\hat{I}_\lambda)
\end{equation}

A total of 1000 random directions is used to compute the sliced Wasserstein distances, $\mathcal{L}_{grad}$ and $\mathcal{L}_{hist}$. 

\subsection{Results}\label{subsec:ms_res}

\begin{figure*}
    \centering
    \includegraphics[width=\linewidth]{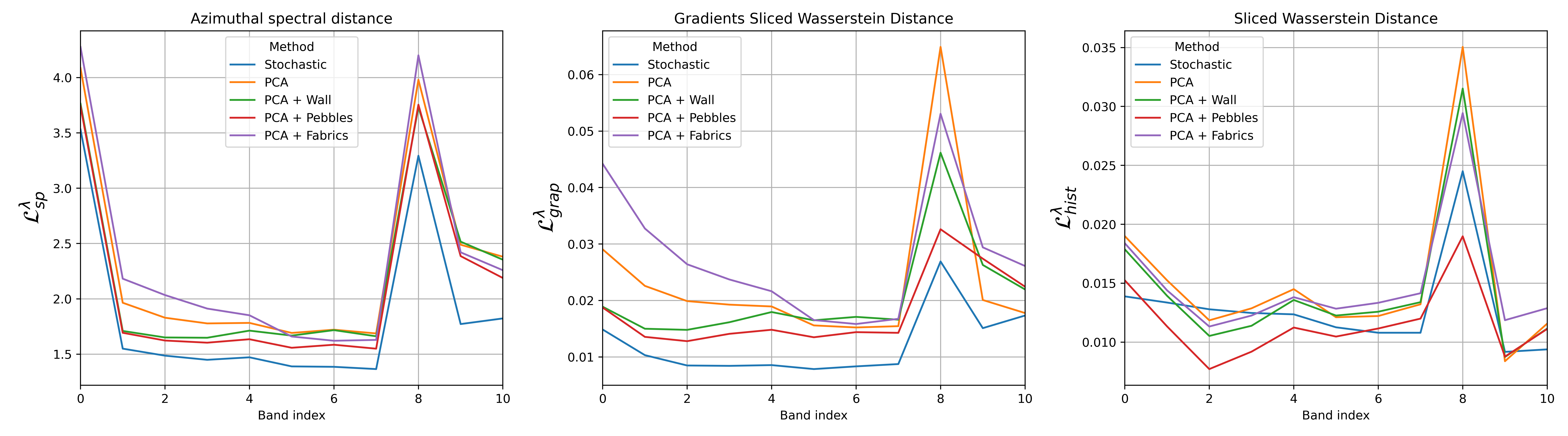}
    \caption{Band wise comparison of the proposed methods.}
    \label{fig:ms_graph_band}
\end{figure*}

We show qualitative results with reconstruction examples in Figure \ref{fig:gatys_ms_rec} and quantitative results in Table \ref{tab:res_gatys_ms_gen}. In addition, we show band-wise comparison for distribution, gradients and spectral metrics in Figure \ref{fig:ms_graph_band}.\\

\subsubsection{Visual quality }
Our methods enable faithful texture reconstruction for cloud field multispectral images. However, results differ depending on the method used. Indeed, we observe more frequently grid-like artifacts on the images synthesized using a projected style distance (either PCA or PCA + Color), while the ones synthesized using the stochastic style distance do not exhibit this kind of artifacts. The PCA method is the one showing the coarsest artifacts and the use of color transfer significantly enhances the visual aspect of the synthetic images.\\

\renewcommand{\h}{0.16\linewidth}
\renewcommand{\hh}{\hspace{0.005\linewidth}}
\renewcommand{\vs}{\vspace{3mm}}
\begin{figure*}
    \centering
    \begin{tikzpicture}[zoomboxarray, zoomboxes below, zoomboxarray columns=1, zoomboxarray rows=1]
        \node [image node] {\includegraphics[width=\h]{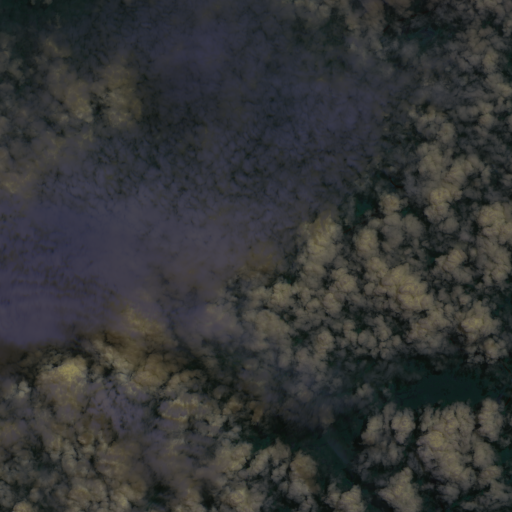}};
    \end{tikzpicture}\hh%
    \begin{tikzpicture}[zoomboxarray, zoomboxes below, zoomboxarray columns=1, zoomboxarray rows=1]
        \node [image node] {\includegraphics[width=\h]{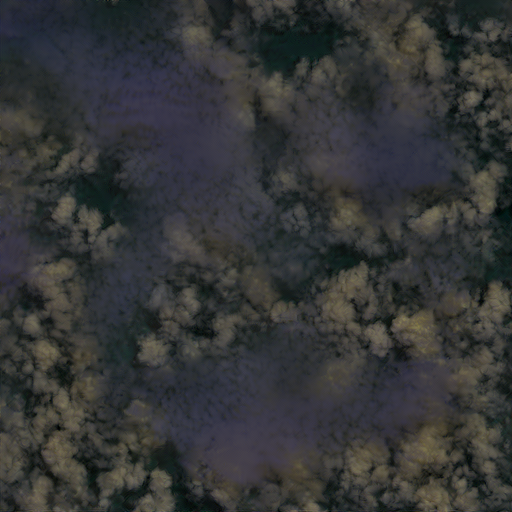}};
        \zoombox[magnification=4]{0.7,0.3}
    \end{tikzpicture}\hh%
    \begin{tikzpicture}[zoomboxarray, zoomboxes below, zoomboxarray columns=1, zoomboxarray rows=1]
        \node [image node] {\includegraphics[width=\h]{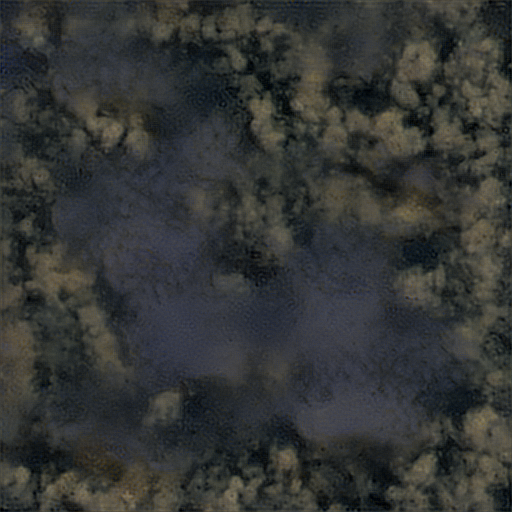}};
        \zoombox[magnification=4]{0.45,0.65}
    \end{tikzpicture}\hh%
    \begin{tikzpicture}[zoomboxarray, zoomboxes below, zoomboxarray columns=1, zoomboxarray rows=1]
        \node [image node] {\includegraphics[width=\h]{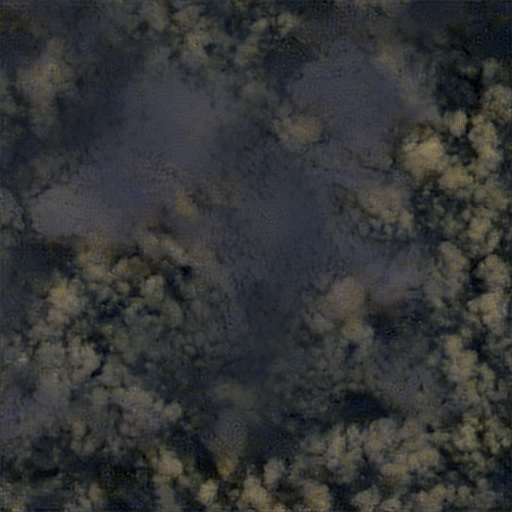}};
        \zoombox[magnification=4]{0.35,0.5}
    \end{tikzpicture}\hh%
    \begin{tikzpicture}[zoomboxarray, zoomboxes below, zoomboxarray columns=1, zoomboxarray rows=1]
        \node [image node] {\includegraphics[width=\h]{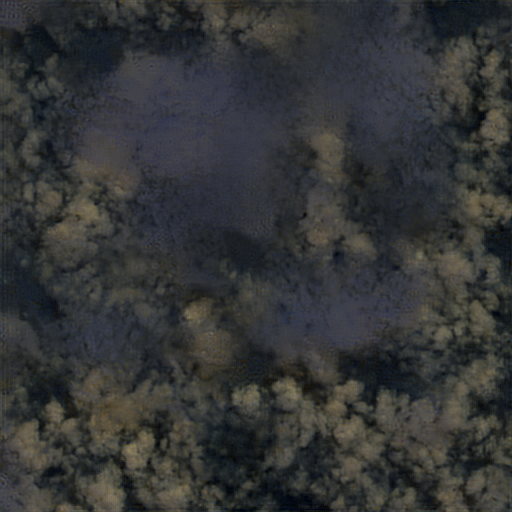}};
        \zoombox[magnification=4]{0.65,0.3}
    \end{tikzpicture}\hh%
    \begin{tikzpicture}[zoomboxarray, zoomboxes below, zoomboxarray columns=1, zoomboxarray rows=1]
        \node [image node] {\includegraphics[width=\h]{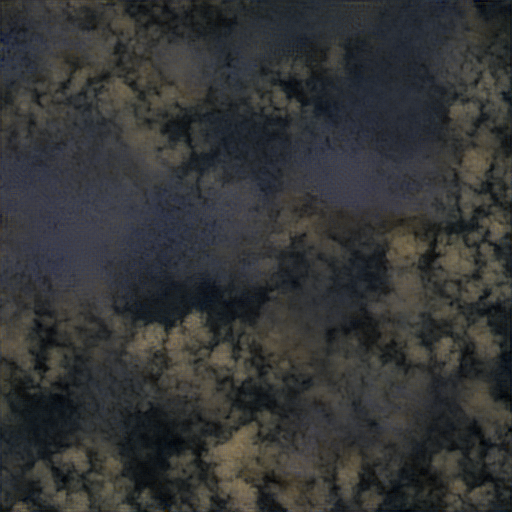}};
        \zoombox[magnification=4]{0.8,0.6}
    \end{tikzpicture}\\
    \vs
    \begin{tikzpicture}[zoomboxarray, zoomboxes below, zoomboxarray columns=1, zoomboxarray rows=1]
        \node [image node] {\includegraphics[width=\h]{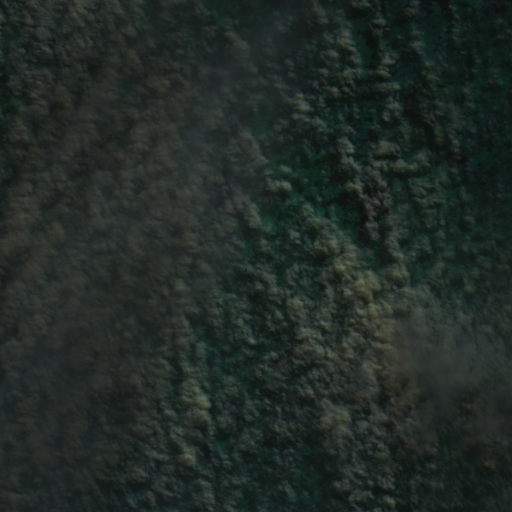}};
    \end{tikzpicture}\hh%
    \begin{tikzpicture}[zoomboxarray, zoomboxes below, zoomboxarray columns=1, zoomboxarray rows=1]
        \node [image node] {\includegraphics[width=\h]{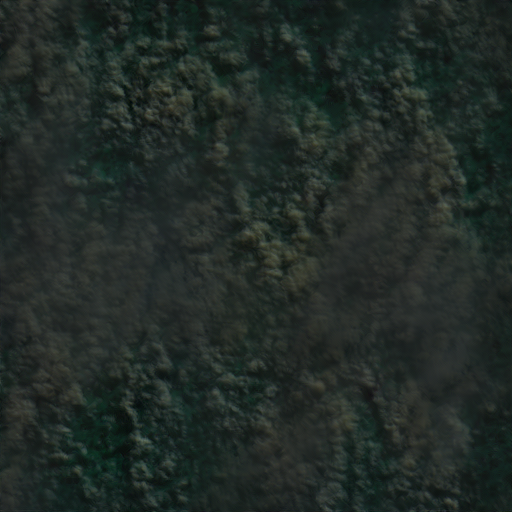}};
        \zoombox[magnification=4]{0.5,0.5}
    \end{tikzpicture}\hh%
    \begin{tikzpicture}[zoomboxarray, zoomboxes below, zoomboxarray columns=1, zoomboxarray rows=1]
        \node [image node] {\includegraphics[width=\h]{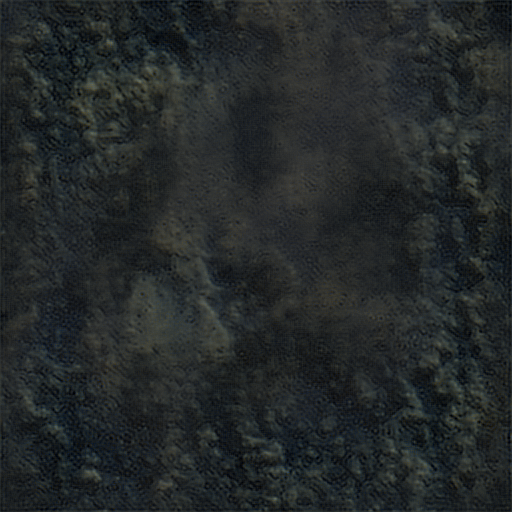}};
        \zoombox[magnification=4]{0.175,0.72}
    \end{tikzpicture}\hh%
    \begin{tikzpicture}[zoomboxarray, zoomboxes below, zoomboxarray columns=1, zoomboxarray rows=1]
        \node [image node] {\includegraphics[width=\h]{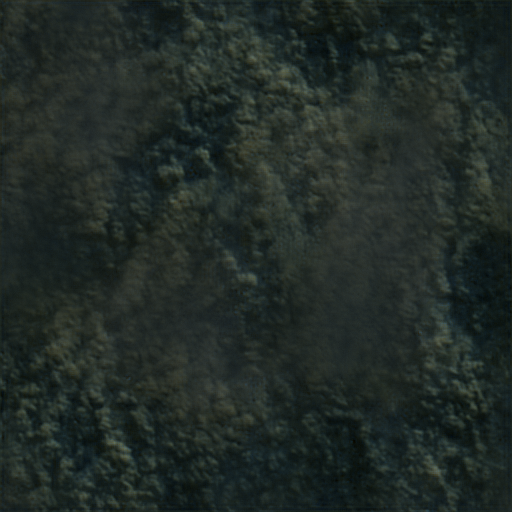}};
        \zoombox[magnification=4]{0.2,0.72}
    \end{tikzpicture}\hh%
    \begin{tikzpicture}[zoomboxarray, zoomboxes below, zoomboxarray columns=1, zoomboxarray rows=1]
        \node [image node] {\includegraphics[width=\h]{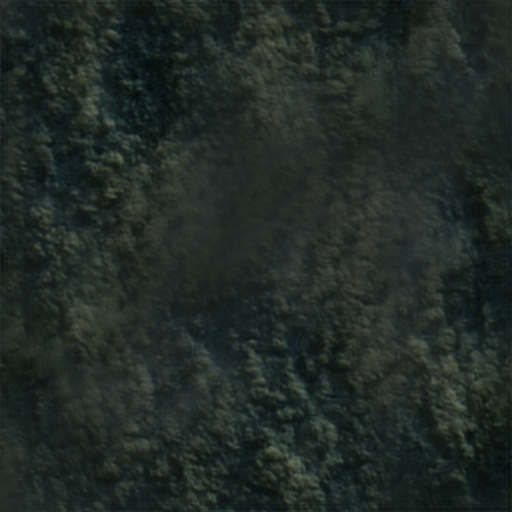}};
        \zoombox[magnification=4]{0.5,0.25}
    \end{tikzpicture}\hh%
    \begin{tikzpicture}[zoomboxarray, zoomboxes below, zoomboxarray columns=1, zoomboxarray rows=1]
        \node [image node] {\includegraphics[width=\h]{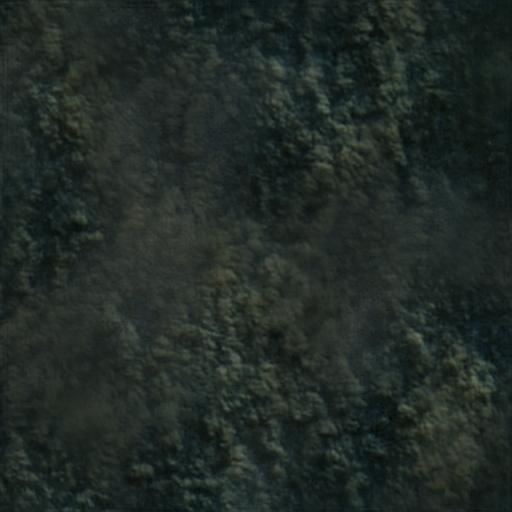}};
        \zoombox[magnification=4]{0.6,0.72}
    \end{tikzpicture}\\
    \vs
    \begin{tikzpicture}[zoomboxarray, zoomboxes below, zoomboxarray columns=1, zoomboxarray rows=1]
        \node [image node] {\includegraphics[width=\h]{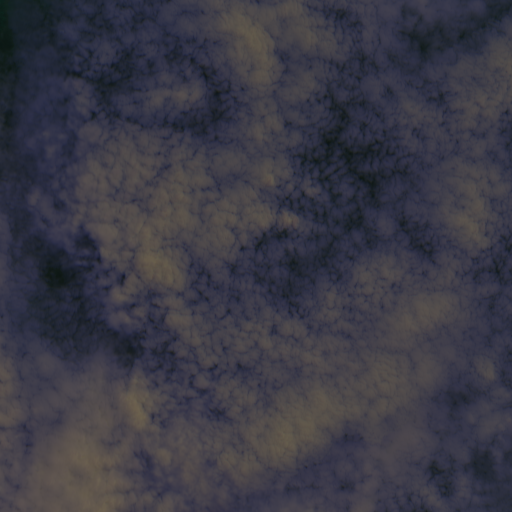}};
    \end{tikzpicture}\hh%
    \begin{tikzpicture}[zoomboxarray, zoomboxes below, zoomboxarray columns=1, zoomboxarray rows=1]
        \node [image node] {\includegraphics[width=\h]{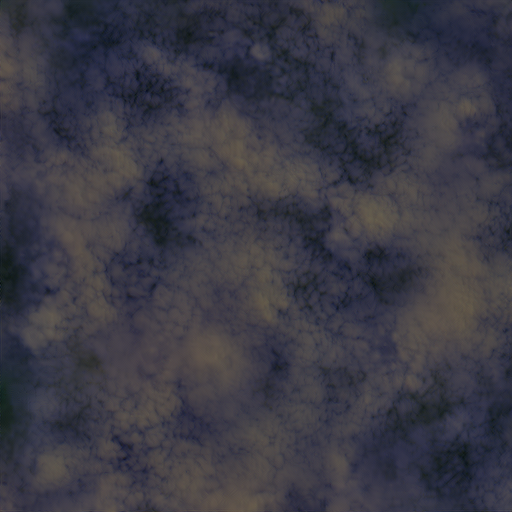}};
        \zoombox[magnification=4]{0.175,0.72}
    \end{tikzpicture}\hh%
    \begin{tikzpicture}[zoomboxarray, zoomboxes below, zoomboxarray columns=1, zoomboxarray rows=1]
        \node [image node] {\includegraphics[width=\h]{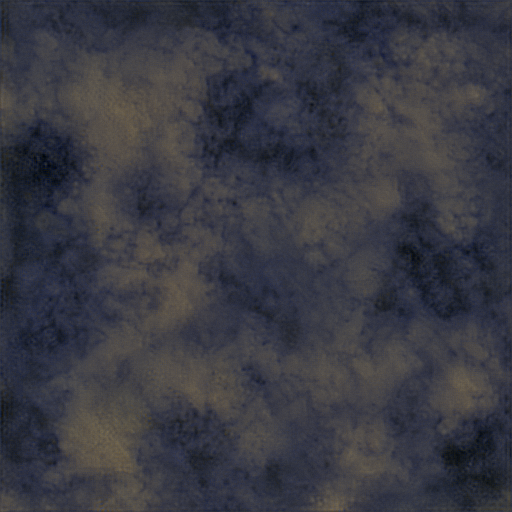}};
        \zoombox[magnification=4]{0.2,0.2}
    \end{tikzpicture}\hh%
    \begin{tikzpicture}[zoomboxarray, zoomboxes below, zoomboxarray columns=1, zoomboxarray rows=1]
        \node [image node] {\includegraphics[width=\h]{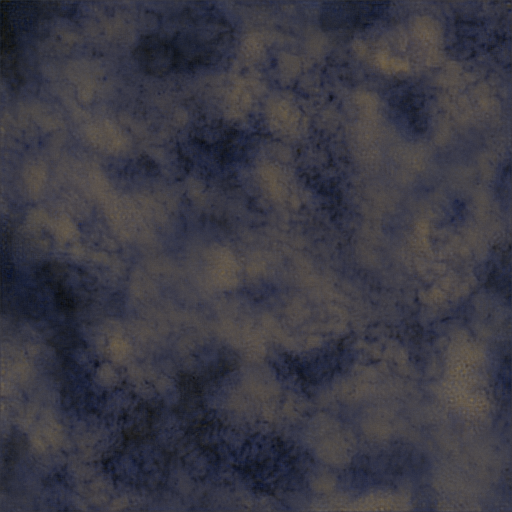}};
        \zoombox[magnification=4]{0.7,0.65}
    \end{tikzpicture}\hh%
    \begin{tikzpicture}[zoomboxarray, zoomboxes below, zoomboxarray columns=1, zoomboxarray rows=1]
        \node [image node] {\includegraphics[width=\h]{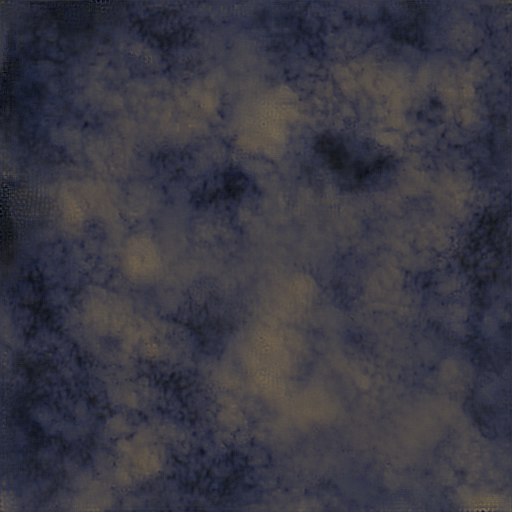}};
        \zoombox[magnification=4]{0.5,0.3}
    \end{tikzpicture}\hh%
    \begin{tikzpicture}[zoomboxarray, zoomboxes below, zoomboxarray columns=1, zoomboxarray rows=1]
        \node [image node] {\includegraphics[width=\h]{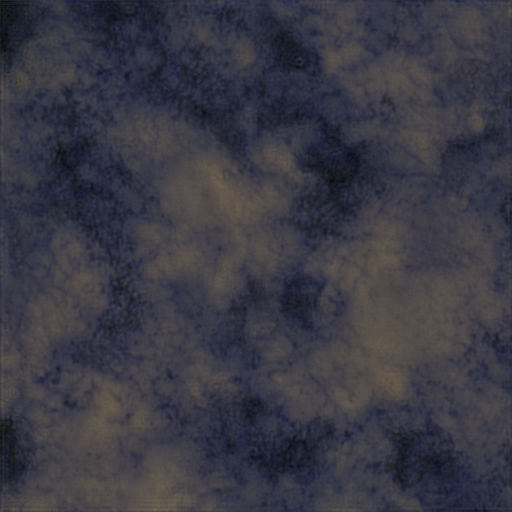}};
        \zoombox[magnification=4]{0.65,0.55}
    \end{tikzpicture}\\  
    \vspace{.2mm}
    \small{%
        \makebox[\h][c]{GT}\hh%
        \makebox[\h][c]{Stochastic}\hh%
        \makebox[\h][c]{PCA}\hh%
        \makebox[\h][c]{PCA + Wall}\hh%
        \makebox[\h][c]{PCA + Pebbles}\hh%
        \makebox[\h][c]{PCA + Fabrics}\\
    }
    \caption{Examples of multispectral texture synthesis using different style distances as optimization objective. The images displayed are obtained by pooling the 11 spectral bands to obtain 3-channels images allowing visual representation of multispectral images (channel 1: bands 1,2,3,4; channel 2: bands 5,6,7,8; channel 3: bands 9,11,12). Exemplar textures are shown on the left, and each following column correspond to a previously presented method of synthesis, label are shown on the bottom. Images that optimize the stochastic loss allow a visual appearance more faithful to the original textures than the ones minimizing a projected style distance, with no grid-like artifacts observed in the first ones. The introduction of color transfer upstream of a projection enables far less of these defects in the generated images.}
    \label{fig:gatys_ms_rec}
\end{figure*}

{
\setlength{\tabcolsep}{4pt}
\begin{table}
    \centering
    \scriptsize{\begin{tabular}{ccccc}
        \toprule
         Method & $\mathcal{L}_{style}^{RGB} \downarrow$ &  $\mathcal{L}_{sp}^{mean} \downarrow$ & $\mathcal{L}_{grad}~(\cdot10^{-3})\downarrow$ & $\mathcal{L}_{hist}~(\cdot10^{-3})\downarrow$\\
        \midrule
        RGB & \textbf{0.019} & 2.03 & 28.53 & 14.09 \\
        \midrule
        Stochastic & 0.099 & \textbf{1.68} & \textbf{9.33} & 12.41 \\
        PCA & 85.15 & 2.09 & 20.55 & 13.02 \\
        PCA + Wall & 55.65 & 1.90 & 15.27 & 11.75 \\
        PCA + Pebbles & 44.71 & 1.87 & 13.46 & \textbf{9.32} \\
        PCA + Fabrics & 95.68 & 2.29 & 27.73 & 12.35 \\
        \bottomrule
    \end{tabular}}
    \caption{Evaluation of our methods on RGB bands. The line RGB refers to a baseline where images are obtained by optimizing only on the 3 RGB bands to match deep features statistics. All metrics are computed on the RGB bands.}
    \label{tab:res_gatys_rgb_gen}
\end{table}}

\begin{table*}
    \centering
    \scriptsize{\begin{tabular}{ccccccccc}
        \toprule
         Method & $\mathcal{L}_{style}^{MS} \downarrow$ & $\mathcal{L}_{style}^{PCA} \downarrow$ &  $\mathcal{L}_{sp}^{mean} \downarrow$ & $\mathcal{L}_{grad}~(\cdot10^{-3})\downarrow$ & $\mathcal{L}_{hist}~(\cdot10^{-3})\downarrow$ & $\mathcal{L}_{\mu}(\cdot 10^{-3}) \downarrow$ & $\mathcal{L}_{\Sigma}(\cdot 10^{-3}) \downarrow$ & $\mathcal{L}_{RX}(\cdot 10^{-3}) \downarrow$ \\
        \midrule
        Stochastic & \textbf{0.17} & 676.84 & \textbf{1.96} & \textbf{13.52} & 12.63 & 17.36 & \textbf{14.02} & 18.62 \\
        PCA & 115.38 & \textbf{0.59} & 2.33 & 26.52 & 15.02 & 12.20 & 21.45 & 15.68 \\
        PCA + Wall & 98.38 & 2158.46 & 2.35 & 21.92 & 14.39 & 9.24 & 21.63 & 12.65 \\
        PCA + Pebbles & 68.00 & 4872.01 & 2.21 & 18.94 & \textbf{11.70} & \textbf{4.58} & 21.01 & \textbf{7.87} \\
        PCA + Fabrics & 102.09 & 11008.26 & 2.33 & 29.78 & 15.32 & 12.80 & 22.16 & 16.83 \\
        \bottomrule
    \end{tabular}}
    \caption{Evaluation of multispectral texture synthesis through CNN trained on RGB images. Metrics were computed and averaged over 130 multispectral images.}
    \label{tab:res_gatys_ms_gen}
\end{table*}

\subsubsection{Comparison with RGB texture synthesis}  

As a sanity check, we compare our method with the result of applying the original method from \cite{gatys_texture_2015} to the three multispectral bands that are closer to the RGB spectral content. For this, we consider bands 2, 3 and 4. 
We perform texture synthesis only using these 3 bands. Then, we compare the obtained 3-channels images with the images obtained by keeping only the same 3 bands after optimizing the multispectral images using the aforementioned multispectral style distances. The quantitative comparison on the 3 RGB bands is given in Table \ref{tab:res_gatys_rgb_gen}. The considered metrics are computed only on the RGB bands. 

Without surprise, the $\mathcal{L}_{style}^{RGB}$ loss is smaller when optimizing the RGB image. Also, the stochastic style distance performs much better than the projected ones for this metric.  Next, one observes that all proposed style distances yield a better or similar preservation of the gradient, the histogram and the spectrum. This may be due to a better account of inter-band dependency when optimizing over all bands. \\

\subsubsection{Comparison between methods } We now use all metrics introduced above to compare the three methods that we have introduced to perform neural multispectral texture synthesis: by using the stochastic style distance, by using the projected distance, and finally by using the projected style distance together with color transfer. Of course, each of these methods perform best when using the metric that has been optimized by the method, and we therefore focus on the remaining metrics to reach some conclusions. Results are gathered in Table \ref{tab:res_gatys_ms_gen}.

First, we observe that the stochastic style distance enforces greater respect of the Fourier spectra and of the gradients in the generated textures, two features that are related to the spatial structure of the images. We can see in Figure \ref{fig:ms_graph_band} that this is the case across all spectral bands. 
 
For the multispectral distribution, two methods, namely \textit{Stochastic} and \textit{PCA + Pebbles}, stand out, showing competitive SWD $\mathcal{L}_{hist}$. However, these results find different explanations if we look at the different statistics. The projected style distance ensures a better respect of the mean $\mathcal{L}_\mu$ while its stochastic counterpart seems to allow a better respect of the higher order statistics, as seen for the covariance $\mathcal{L}_\Sigma$. Then, we also observe different dynamics across the spectral bands with the \textit{Stochastic} method outperforming \textit{PCA + Pebbles} for 4 out of 11 spectral bands.

Finally, the band wise analysis clearly reveals that all methods under-perform on the 2 correction bands, at 442nm and 945nm (see Figure \ref{fig:ms_graph_band}). This was expected as they present more variability than other bands. However, the two most effective methods are less sensitive to this high variability, showing greater ability to take multispectral information into account. 

\section{Conclusion}

In this paper, we have introduced several strategies to generalize state-of-the-art RGB texture synthesis methods to multispectral images. The main difficulty to do so is to generalize style distances obtained from CNN pre-trained on 3-channels images to a larger number of bands.  

The new introduced style distances were used to perform exemplar-based texture synthesis on single multispectral images of cloud fields from the Sentinel-2 mission. They show satisfactory results, both visually and numerically.

The stochastic style distance ensures the best visual reconstruction and globally shows the best numerical results among the tested methods, but comes at a higher computational cost. The use of a projection is faster but less efficient. However, color transfer upstream of the projection happens to largely enhance the quality of the reconstruction.

Finally, all of the proposed methods do not require any additional training, since they leverage an RGB CNN that has proven itself to perform well for texture synthesis and style transfer. The solutions we propose are not specific to texture modeling and can be applied to a wide range of style-like distances that take advantage of the capabilities of pre-trained neural networks. The most direct perspective we can highlight is style transfer. A second perspective is the use of the proposed style distances for feed-forward approaches, to texture synthesis or to more general restoration or editing tasks. Eventually, common distances based on neural networks such as the widely used FID \cite{heusel_gans_2017} or LPIPS \cite{zhang_unreasonable_2018} can also be extended to multi/hyperspectral imaging with the strategies proposed in this paper, opening interesting avenues for multispectral image quality evaluation. Another perspective lies in the use of the texture models presented here to train new networks to perform multi-texture synthesis.

\appendices
\renewcommand{\h}{0.85\linewidth}
\begin{figure*}
    \centering
    \includegraphics[width=\h]{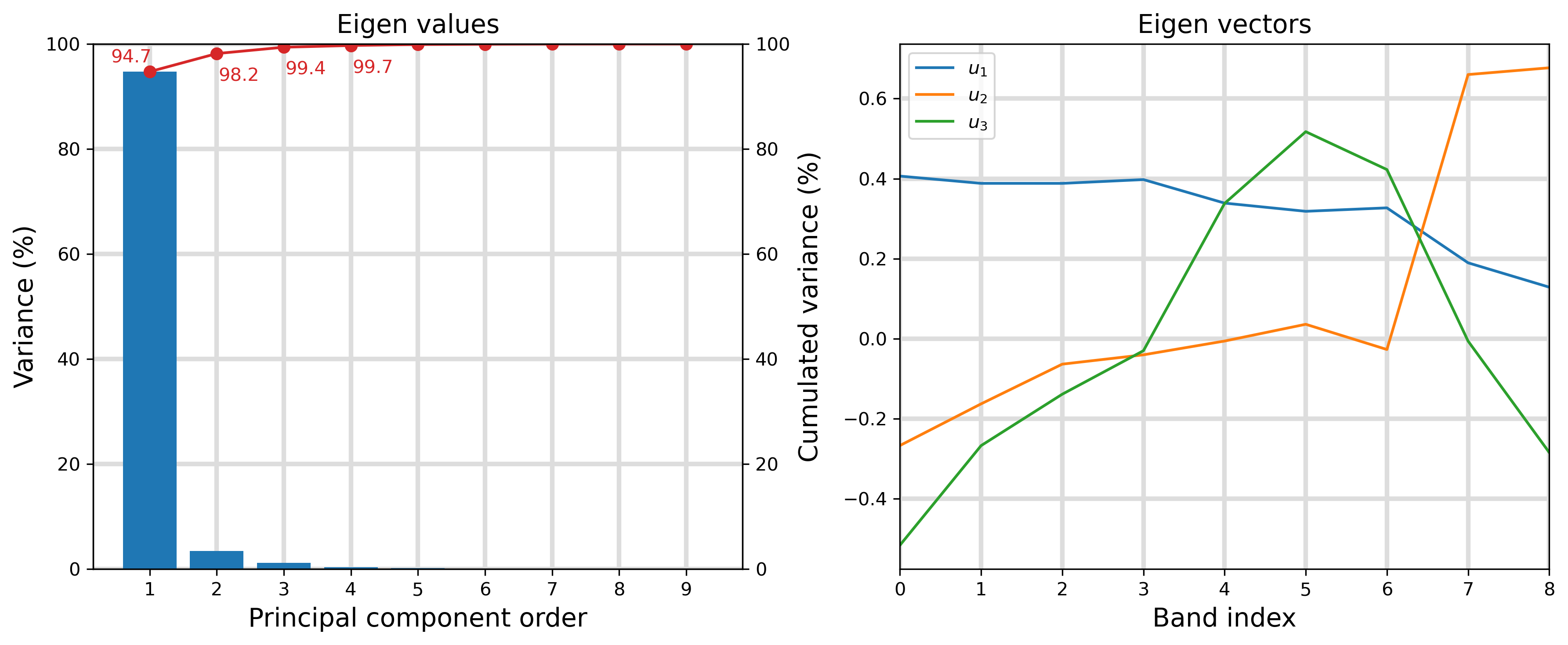}
    \caption{(Left) Eigenvalues of the used PCA over 9-band Sentinel-2 images. The first component encompasses $95\%$ of the variance, the 3 first $99\%$. (Right) 3 first eigenvectors of the used PCA over 9-band Sentinel-2 images (denoted as $u_i$).}
    \label{fig:pca_9b}
\end{figure*}

\begin{table*}
    \centering
    \scriptsize{\begin{tabular}{ccccccccc}
        \toprule
         Method & $\mathcal{L}_{style}^{MS} \downarrow$ & $\mathcal{L}_{style}^{PCA} \downarrow$ &  $\mathcal{L}_{sp}^{mean} \downarrow$ & $\mathcal{L}_{grad}~(\cdot 10^{-3})$ & $\mathcal{L}_{hist}~(\cdot 10^{-3}) \downarrow$ & $\mathcal{L}_{\mu}(\cdot 10^{-3}) \downarrow$ & $\mathcal{L}_{\Sigma}(\cdot 10^{-3}) \downarrow$ & $\mathcal{L}_{RX}(\cdot 10^{-3}) \downarrow$ \\ \midrule
        Stochastic & \textbf{0.09} & \textbf{161.51} & \textbf{1.72} & \textbf{12.46} & 9.19 & 14.69 & \textbf{4.93} & 15.27 \\
        PCA + Pebbles & 27.92 & 745.75 & 1.88 & 16.30 & \textbf{7.66} & \textbf{3.10} & 7.08 & 4.04 \\
        \bottomrule
    \end{tabular}}
    \caption{Mean principal metrics results for multispectral texture synthesis on 9-band Sentinel-2 images. Sliced distances $\mathcal{L}_{sp}^{mean}$ and $\mathcal{L}_{hist}$ were computed using 1000 random directions.}
    \label{tab:res_gatys_ms_9b_gen}
\end{table*}

\section{9 bands dataset}\label{appendix:9bands}

{
\setlength{\tabcolsep}{2.5pt}
\begin{table}
    \centering
    \scriptsize{\begin{tabular}{cccccc}
        \toprule
         $\begin{array}{c}
    \text{Number} \\
    \text{of bands}
\end{array}$ & Method & $\mathcal{L}_{style}^{RGB} \downarrow$ &  $\mathcal{L}_{sp}^{mean} \downarrow$ &  $\begin{array}{c}
    \mathcal{L}_{grad} \\
    (\cdot 10^{-3})
\end{array}\big\downarrow$ & $\begin{array}{c}
    \mathcal{L}_{hist} \\
    (\cdot 10^{-3})
\end{array}\big\downarrow$\\
        \midrule
        3 & RGB & \textbf{0.019} & 2.03 & 3.09 & 1.56 \\
        \midrule
        \multirow{2}{*}{9} & Stochastic & 0.059 & \textbf{1.58} & \textbf{9.32} & 9.75 \\
         & PCA + Pebbles & 31.15 & 1.68 & 11.63 & \textbf{8.25} \\
        \midrule
        \multirow{2}{*}{11} & Stochastic & 0.099 & 1.68 & 9.33 & 12.41 \\
         & PCA + Pebbles & 44.71 & 1.87 & 13.46 & 9.32 \\
        \bottomrule
    \end{tabular}}
    \caption{Mean principal metrics results computed on RGB bands. The line RGB refers to the benchmark where the 3 bands images were optimized to match deep features statistics}
    \label{tab:res_gatys_9b_rgb_gen}
\end{table}}

In addition to the results presented for the 11-band Sentinel-2 images, we present in this section the results of texture synthesis for 9-band images. These images correspond to the same images keeping only the proper observation bands and excluding the 2 spectral bands that are usually used for atmospheric correction (see \ref{sec:data}). We chose to test this new dataset only on the 2 best performing methods, namely \textit{Stochastic} and \textit{PCA + Pebbles}. Figure \ref{fig:pca_9b} presents the eigenvectors and eigenvalues of PCA used for the 9-band dataset.

Numerical results are shown in Table \ref{tab:res_gatys_ms_9b_gen}. These results are in line with the previous results on the 11-band images with regard to the hierarchy of the two methods in the various metrics. Moreover, as the two bands that presented the more variability are removed, the synthesis task becomes easier. Indeed, those bands shown themselves to be harder to synthesize for the 11-band images and the results on the 9-band images show a quantitative improvement for both of the tested methods. The same goes when we compare the synthesis on the 3 RGB bands with the benchmark (consisting of optimizing the images only on these 3 bands) and the 11-band dataset: a fewer number of bands to consider enhance the quality of the reconstruction for both tested methods.

\section{Stochastic style distance batch size}\label{appendix:sto_batch_size}  

\begin{figure*}
    \centering
    \includegraphics[width=\linewidth]{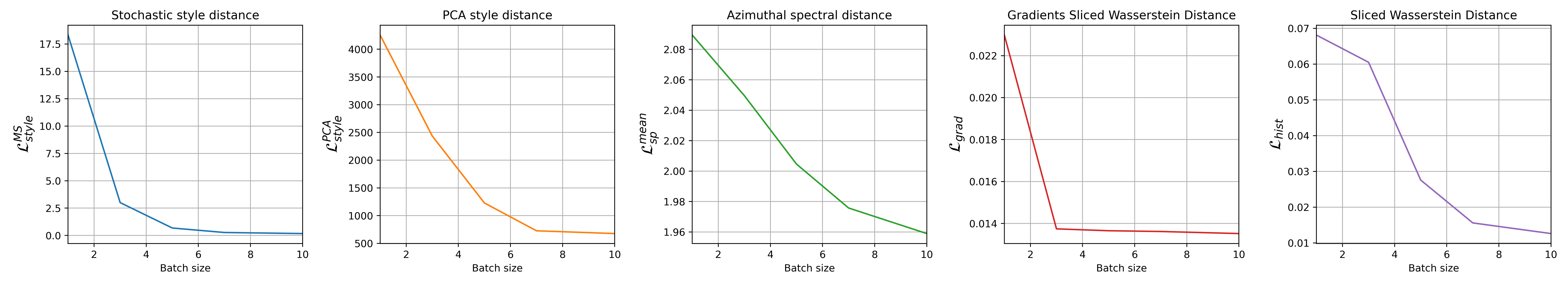}
    \caption{Influence of the number of triplets drawn at each iteration on texture synthesis quality. Metrics are averaged on 130 images.}
    \label{fig:ms_graph_sto_bs}
\end{figure*}

\renewcommand{\h}{0.16\linewidth}
\renewcommand{\hh}{\hspace{0.005\linewidth}}
\renewcommand{\vs}{\vspace{1.5mm}}
\begin{figure*}
    \centering
    \includegraphics[width=\h]{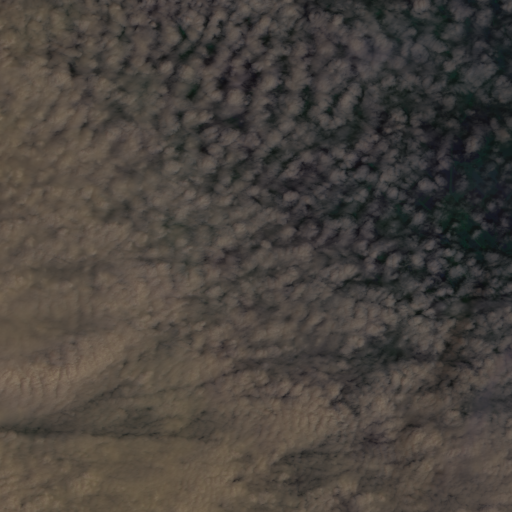}
    \includegraphics[width=\h]{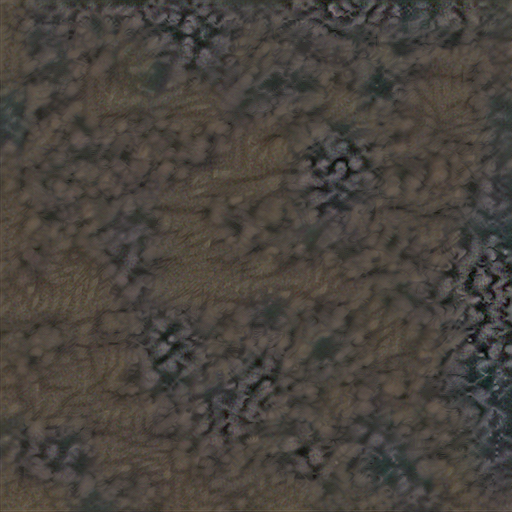}
    \includegraphics[width=\h]{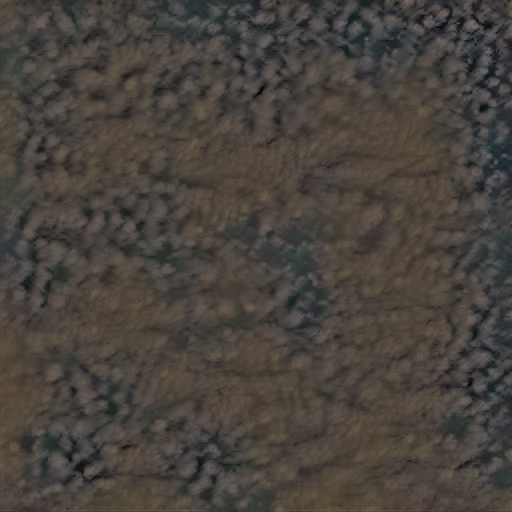}
    \includegraphics[width=\h]{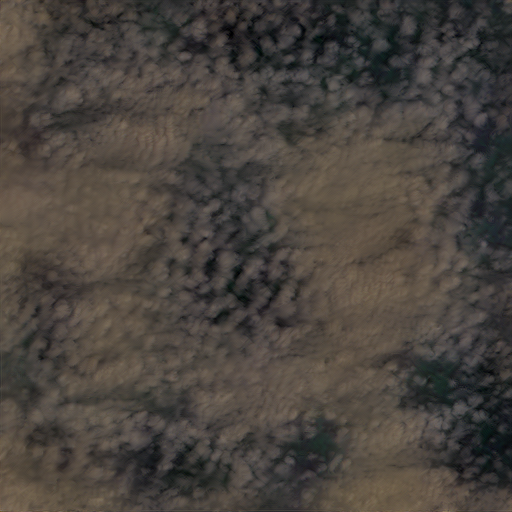}
    \includegraphics[width=\h]{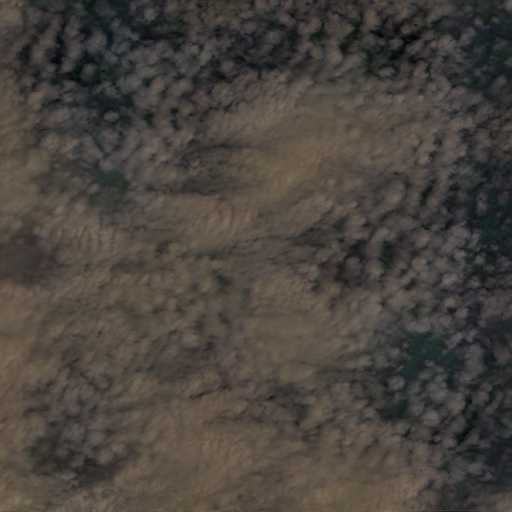}
    \includegraphics[width=\h]{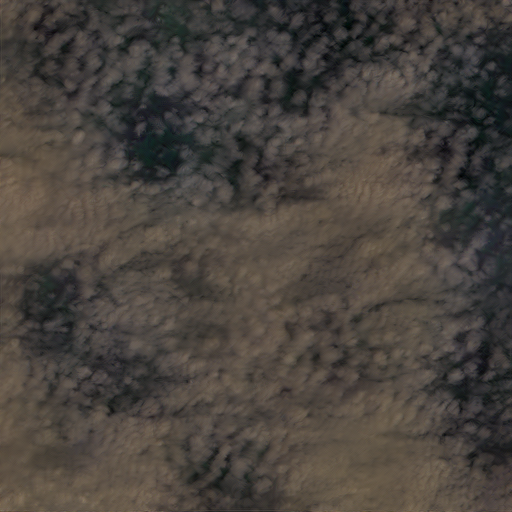}\\
    \vs
    \includegraphics[width=\h]{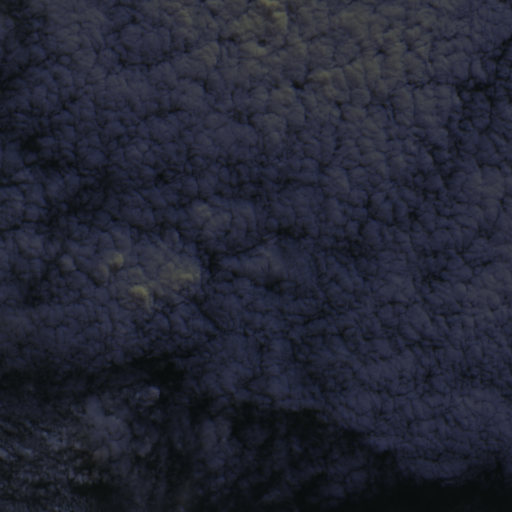}
    \includegraphics[width=\h]{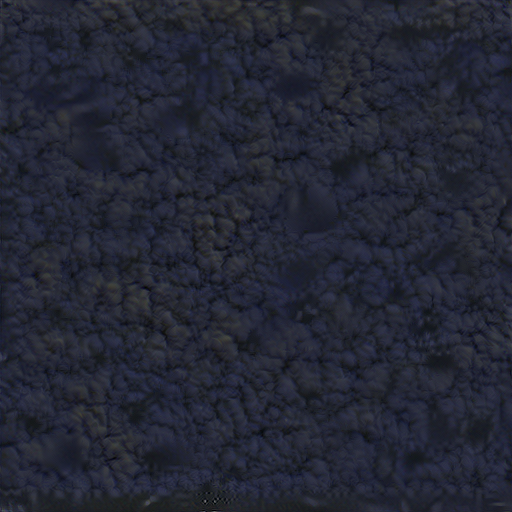}
    \includegraphics[width=\h]{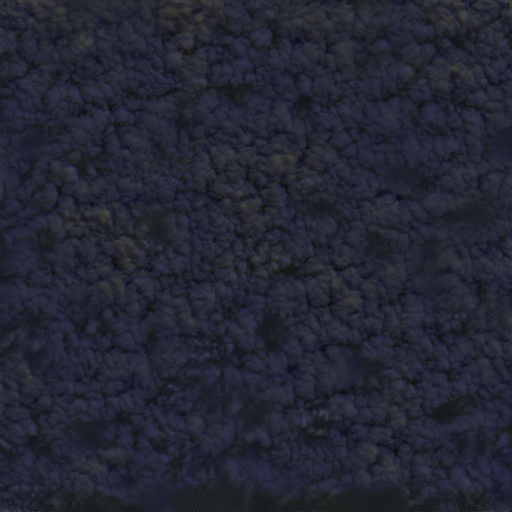}
    \includegraphics[width=\h]{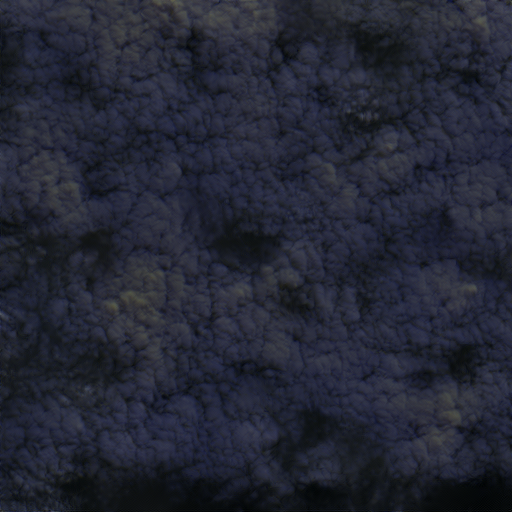}
    \includegraphics[width=\h]{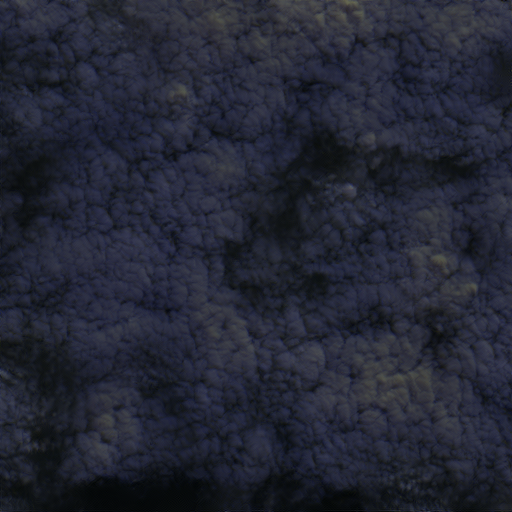}
    \includegraphics[width=\h]{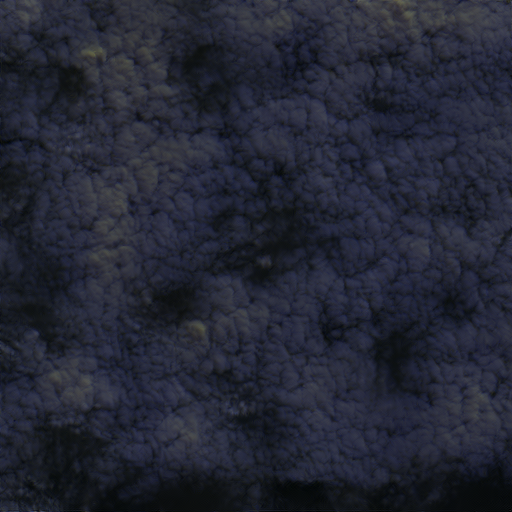}\\
    \vspace{.5mm}
    \small{%
        \makebox[\h][c]{GT}\hh%
        \makebox[\h][c]{$N_B = 1$}\hh%
        \makebox[\h][c]{$N_B = 3$}\hh%
        \makebox[\h][c]{$N_B = 5$}\hh%
        \makebox[\h][c]{$N_B = 7$}\hh%
        \makebox[\h][c]{$N_B = 10$}\\
    }
    \caption{Examples of multispectral texture reconstruction using stochastic style distance as minimization objective and varying the number of triplets $N_B$ drawn at each iterations. The images displayed are obtained by pooling the 11 spectral bands to obtain 3-channels (see \ref{fig:gatys_ms_rec} for details). Exemplar textures are shown on the left, and each following column correspond to a batch size, label are shown on the bottom.}
    \label{fig:gatys_ms_rec_sto_bs}
\end{figure*}

In this section we study the influence of the batch size (i.e. the number of triplets drawn at each iteration) on the synthesis results when using the stochastic style distance as a minimization objective. We performed exemplar synthesis on the same 130 images from the Sentinel-2 mission and vary the batch size. 

{
\setlength{\tabcolsep}{4.3pt}
\begin{table}
    \centering
    \scriptsize{\begin{tabular}{cccccc}
        \toprule
        Batch size & $\mathcal{L}_{style}^{MS} \downarrow$ & $\mathcal{L}_{style}^{PCA} \downarrow$ &  $\mathcal{L}_{sp}^{mean} \downarrow$ & $\begin{array}{c}
    \mathcal{L}_{grad} \\
    (\cdot 10^{-3})
\end{array}\big\downarrow$ & $\begin{array}{c}
    \mathcal{L}_{hist} \\
    (\cdot 10^{-3})
\end{array}\big\downarrow$\\
        \midrule
        1 & 18.32 & 4254.55 & 2.09 & 22.98 & 68.11 \\
        3 & 3.00 & 2435.40 & 2.05 & 13.74 & 60.51 \\
        5 & 0.68 & 1228.71 & 2.01 & 13.65 & 27.59 \\
        7 & 0.27 & 726.56 & 1.98 & 13.62 & 15.60 \\
        10 & \textbf{0.17} & \textbf{676.84} & \textbf{1.96} & \textbf{13.52} & \textbf{12.63} \\
        \bottomrule
    \end{tabular}}
    \caption{Mean principal metrics results for multispectral texture synthesis on 11-band Sentinel-2 images when varying the the number of triplets drawn at each iteration}
    \label{tab:res_gatys_ms_bs_gen}
\end{table}}

The numerical results of our experiment are given in Table \ref{tab:res_gatys_ms_bs_gen} and illustrated in Figure \ref{fig:ms_graph_sto_bs}. We also provide visual examples in Figure \ref{fig:gatys_ms_rec_sto_bs}. They show that the higher the batch size is, the better are the results for all the tested metric. We observe a great interest in increasing the batch size up to at least 5 triplets. Indeed, we also observe this threshold in visual quality starting at 5 images for the majority of the tested images.

\section*{Acknowledgment}
This work was supported by the Defence Innovation Agency.

\bibliographystyle{plain}
\bibliography{references.bib}

\begin{IEEEbiography}[{\includegraphics[width=1in, height=1.25in, clip, keepaspectratio]{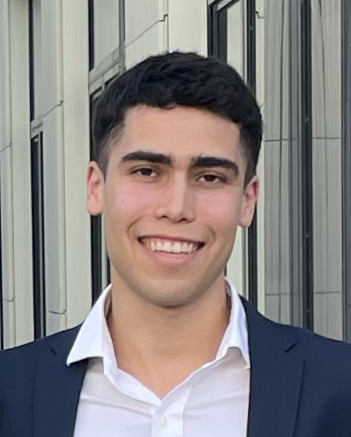}}]{Sélim Ollivier}
    received the M.Eng in mathematics and data science and the M.Sc in fundamental and applied mathematics from CentraleSupélec, Gif-sur-Yvette, France in 2023. He is currently pursuing the Ph.D degree with the LTCI, Télécom Paris, Palaiseau, France and with the DOTA/AILab, ONERA, Palaiseau, France, under the supervision of Yann Gousseau and Sidonie Lefebvre.
    He is working on the development of deep learning models for texture synthesis applied to multi/hyperspectral and 3D imaging.
\end{IEEEbiography}

\begin{IEEEbiography}[{\includegraphics[width=1in, height=1.25in, clip, keepaspectratio]{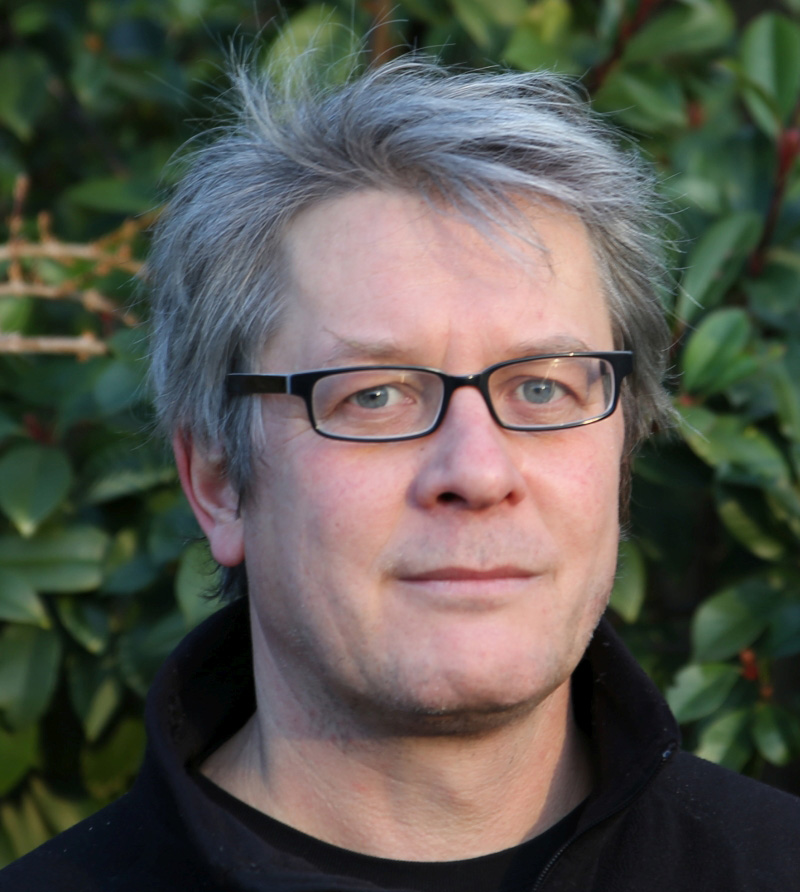}}]{Yann Gousseau}
    Yann Gousseau received the engineering degree from the École Centrale de Paris, France, the Part III of the Mathematical Tripos from the University of Cambridge, UK, and the Ph.D. degree in Applied Mathematics from the University of Paris-Dauphine. He is currently a professor at Telecom Paris, IP Paris. His research interests include the mathematical modeling of natural images and textures, generative models, computer vision, image and video processing.
\end{IEEEbiography}

\begin{IEEEbiography}[{\includegraphics[width=1in, height=1.25in, clip, keepaspectratio]{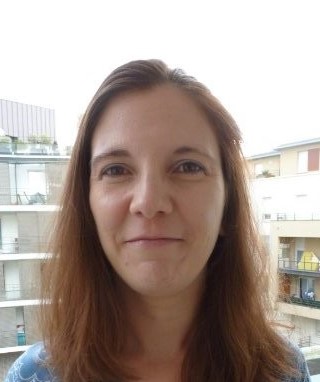}}]{Sidonie Lefebvre}
 received the Ph.D. degree from Ecole Centrale Paris in 2006. She is
currently working as a Research Scientist in statistics at ONERA Palaiseau (France), in
the Optics and Associated Techniques Department. Her research interests are centered
on the design and modeling of computer experiments, uncertainty quantification and
sensitivity analysis, and target detection.   
\end{IEEEbiography}
\end{document}

%% file: zoomboxes.tex
\usetikzlibrary{spy,calc}

\newif\ifblackandwhitecycle
\gdef\patternnumber{0}

\pgfkeys{/tikz/.cd,
    zoombox paths/.style={
        draw=orange,
        very thick
    },
    black and white/.is choice,
    black and white/.default=static,
    black and white/static/.style={ 
        draw=white,   
        zoombox paths/.append style={
            draw=white,
            postaction={
                draw=black,
                loosely dashed
            }
        }
    },
    black and white/static/.code={
        \gdef\patternnumber{1}
    }, n
    black and white/cycle/.code={
        \blackandwhitecycletrue
        \gdef\patternnumber{1}
    },
    black and white pattern/.is choice,
    black and white pattern/0/.style={},
    black and white pattern/1/.style={    
            draw=white,
            postaction={
                draw=black,
                dash pattern=on 2pt off 2pt
            }
    },
    black and white pattern/2/.style={    
            draw=white,
            postaction={
                draw=black,
                dash pattern=on 4pt off 4pt
            }
    },
    black and white pattern/3/.style={    
            draw=white,
            postaction={
                draw=black,
                dash pattern=on 4pt off 4pt on 1pt off 4pt
            }
    },
    black and white pattern/4/.style={    
            draw=white,
            postaction={
                draw=black,
                dash pattern=on 4pt off 2pt on 2 pt off 2pt on 2 pt off 2pt
            }
    },
    zoomboxarray inner gap/.initial=5pt,
    zoomboxarray columns/.initial=2,
    zoomboxarray rows/.initial=2,
    subfigurename/.initial={},
    figurename/.initial={zoombox},
    zoomboxarray/.style={
        execute at begin picture={
            \begin{scope}[
                spy using outlines={%
                    zoombox paths,
                    width=\imagewidth / \pgfkeysvalueof{/tikz/zoomboxarray columns} - (\pgfkeysvalueof{/tikz/zoomboxarray columns} - 1) / \pgfkeysvalueof{/tikz/zoomboxarray columns} * \pgfkeysvalueof{/tikz/zoomboxarray inner gap} -\pgflinewidth,
                    height=\imageheight / \pgfkeysvalueof{/tikz/zoomboxarray rows} - (\pgfkeysvalueof{/tikz/zoomboxarray rows} - 1) / \pgfkeysvalueof{/tikz/zoomboxarray rows} * \pgfkeysvalueof{/tikz/zoomboxarray inner gap}-\pgflinewidth,
                    magnification=3,
                    every spy on node/.style={
                        zoombox paths
                    },
                    every spy in node/.style={
                        zoombox paths
                    }
                }
            ]
        },
        execute at end picture={
            \end{scope}
        },
        spymargin/.initial=0.5em,
        zoomboxes xshift/.initial=1,
        zoomboxes right/.code=\pgfkeys{/tikz/zoomboxes xshift=1},
        zoomboxes left/.code=\pgfkeys{/tikz/zoomboxes xshift=-1},
        zoomboxes yshift/.initial=0,
        zoomboxes above/.code={
            \pgfkeys{/tikz/zoomboxes yshift=1},
            \pgfkeys{/tikz/zoomboxes xshift=0}
        },
        zoomboxes below/.code={
            \pgfkeys{/tikz/zoomboxes yshift=-1},
            \pgfkeys{/tikz/zoomboxes xshift=0}
        },
        caption margin/.initial=0.5ex,
    },
    adjust caption spacing/.code={},
    image container/.style={
        inner sep=0pt,
        at=(image.north),
        anchor=north,
        adjust caption spacing
    },
    zoomboxes container/.style={
        inner sep=0pt,
        at=(image.north),
        anchor=north,
        name=zoomboxes container,
        xshift=\pgfkeysvalueof{/tikz/zoomboxes xshift}*(\imagewidth+\pgfkeysvalueof{/tikz/spymargin}),
        yshift=\pgfkeysvalueof{/tikz/zoomboxes yshift}*(\imageheight+\pgfkeysvalueof{/tikz/spymargin}+\pgfkeysvalueof{/tikz/caption margin}),
        adjust caption spacing
    },
    calculate dimensions/.code={
        \pgfpointdiff{\pgfpointanchor{image}{south west} }{\pgfpointanchor{image}{north east} }
        \pgfgetlastxy{\imagewidth}{\imageheight}
        \global\let\imagewidth=\imagewidth
        \global\let\imageheight=\imageheight
        \gdef\columncount{1}
        \gdef\rowcount{1}
        
    },
    image node/.style={
        inner sep=0pt,
        name=image,
        anchor=south west,
        append after command={
            [calculate dimensions]
            node [image container,subfigurename=\pgfkeysvalueof{/tikz/figurename}-image] {\phantomimage}
            node [zoomboxes container,subfigurename=\pgfkeysvalueof{/tikz/figurename}-zoom] {\phantomimage}
        }
    },
    color code/.style={
        zoombox paths/.append style={draw=#1}
    },
    connect zoomboxes/.style={
    spy connection path={\draw[draw=none,zoombox paths] (tikzspyonnode) -- (tikzspyinnode);}
    },
    help grid code/.code={
        \begin{scope}[
                x={(image.south east)},
                y={(image.north west)},
                font=\footnotesize,
                help lines,
                overlay
            ]
            \foreach \x in {0,1,...,9} { 
                \draw(\x/10,0) -- (\x/10,1);
                \node [anchor=north] at (\x/10,0) {0.\x};
            }
            \foreach \y in {0,1,...,9} {
                \draw(0,\y/10) -- (1,\y/10);                        \node [anchor=east] at (0,\y/10) {0.\y};
            }
        \end{scope}    
    },
    help grid/.style={
        append after command={
            [help grid code]
        }
    },
}

%% file: main.bbl
\begin{thebibliography}{10}

\bibitem{alibani_multispectral_2024}
Michael Alibani, Nicola Acito, and Giovanni Corsini.
\newblock Multispectral {Satellite} {Image} {Generation} {Using} {StyleGAN3}.
\newblock {\em IEEE Journal of Selected Topics in Applied Earth Observations
  and Remote Sensing}, 17:4379--4391, 2024.

\bibitem{chatillon_geometrically_2023}
Pierrick Chatillon, Yann Gousseau, and Sidonie Lefebvre.
\newblock A {Geometrically} {Aware} {Auto}-{Encoder} for {Multi}-texture
  {Synthesis}.
\newblock In Luca Calatroni, Marco Donatelli, Serena Morigi, Marco Prato, and
  Matteo Santacesaria, editors, {\em Scale {Space} and {Variational} {Methods}
  in {Computer} {Vision}}, pages 263--275, Cham, 2023. Springer International
  Publishing.

\bibitem{chu_ghost_2021}
Rui~Jian Chu, Noel Richard, Hermine Chatoux, Christine Fernandez-Maloigne, and
  Jon~Yngve Hardeberg.
\newblock Ghost: {Gradient} {Histogram} of {Spectral} {Texture}.
\newblock In {\em 2021 11th {Workshop} on {Hyperspectral} {Imaging} and
  {Signal} {Processing}: {Evolution} in {Remote} {Sensing} ({WHISPERS})}, pages
  1--5, March 2021.

\bibitem{chu_metrological_2019}
Rui~Jian Chu, Noel Richard, Faouzi Ghorbel, Christine Fernandez-Maloigne, and
  Jon~Yngve Hardeberg.
\newblock A {Metrological} {Framework} {For} {Hyperspectral} {Texture}
  {Analysis} {Using} {Relative} {Spectral} {Difference} {Occurrence} {Matrix}.
\newblock In {\em 2019 10th {Workshop} on {Hyperspectral} {Imaging} and
  {Signal} {Processing}: {Evolution} in {Remote} {Sensing} ({WHISPERS})}, pages
  1--5, September 2019.

\bibitem{chu_hyperspectral_2021}
Rui~Jian Chu, Noël Richard, Hermine Chatoux, Christine Fernandez-Maloigne, and
  Jon~Yngve Hardeberg.
\newblock Hyperspectral {Texture} {Metrology} {Based} on {Joint} {Probability}
  of {Spectral} and {Spatial} {Distribution}.
\newblock {\em IEEE Transactions on Image Processing}, 30:4341--4356, 2021.

\bibitem{cong_satmae_2022}
Yezhen Cong, Samar Khanna, Chenlin Meng, Patrick Liu, Erik Rozi, Yutong He,
  Marshall Burke, David~B. Lobell, and Stefano Ermon.
\newblock {SatMAE}: {Pre}-training {Transformers} for {Temporal} and
  {Multi}-{Spectral} {Satellite} {Imagery}.
\newblock October 2022.
\newblock arXiv 2207.08051.

\bibitem{czerkawski_satellitecloudgenerator_2023}
Mikolaj Czerkawski, Robert Atkinson, Craig Michie, and Christos Tachtatzis.
\newblock {SatelliteCloudGenerator}: {Controllable} {Cloud} and {Shadow}
  {Synthesis} for {Multi}-{Spectral} {Optical} {Satellite} {Images}.
\newblock {\em Remote Sensing}, 15(17):4138, January 2023.
\newblock Number: 17 Publisher: Multidisciplinary Digital Publishing Institute.

\bibitem{gatys_texture_2015}
Leon Gatys, Alexander~S Ecker, and Matthias Bethge.
\newblock Texture {Synthesis} {Using} {Convolutional} {Neural} {Networks}.
\newblock In {\em Advances in {Neural} {Information} {Processing} {Systems}},
  volume~28, 2015.

\bibitem{ge_cloud_2023}
Kaiqiang Ge, Jiayin Liu, Feng Wang, Bo~Chen, and Yuxin Hu.
\newblock A {Cloud} {Detection} {Method} {Based} on {Spectral} and {Gradient}
  {Features} for {SDGSAT}-1 {Multispectral} {Images}.
\newblock {\em Remote Sensing}, 15(1):24, January 2023.
\newblock Number: 1 Publisher: Multidisciplinary Digital Publishing Institute.

\bibitem{heeger_pyramid-based_1995}
D.J. Heeger and J.R. Bergen.
\newblock Pyramid-based texture analysis/synthesis.
\newblock In {\em Proceedings., {International} {Conference} on {Image}
  {Processing}}, volume~3, pages 648--651 vol.3, October 1995.

\bibitem{heusel_gans_2017}
Martin Heusel, Hubert Ramsauer, Thomas Unterthiner, Bernhard Nessler, and Sepp
  Hochreiter.
\newblock {GANs} {Trained} by a {Two} {Time}-{Scale} {Update} {Rule} {Converge}
  to a {Local} {Nash} {Equilibrium}.
\newblock In {\em Advances in {Neural} {Information} {Processing} {Systems}},
  volume~30. Curran Associates, Inc., 2017.

\bibitem{jetchev_texture_2017}
Nikolay Jetchev, Urs Bergmann, and Roland Vollgraf.
\newblock Texture {Synthesis} with {Spatial} {Generative} {Adversarial}
  {Networks}, September 2017.
\newblock arXiv:1611.08207 [cs, stat].

\bibitem{khan_hytexila_2018}
Haris~Ahmad Khan, Sofiane Mihoubi, Benjamin Mathon, Jean-Baptiste Thomas, and
  Jon~Yngve Hardeberg.
\newblock {HyTexiLa}: {High} {Resolution} {Visible} and {Near} {Infrared}
  {Hyperspectral} {Texture} {Images}.
\newblock {\em Sensors}, 18(7):2045, July 2018.
\newblock Number: 7 Publisher: Multidisciplinary Digital Publishing Institute.

\bibitem{khanna_diffusionsat_2024}
Samar Khanna, Patrick Liu, Linqi Zhou, Chenlin Meng, Robin Rombach, Marshall
  Burke, David Lobell, and Stefano Ermon.
\newblock {DiffusionSat}: {A} {Generative} {Foundation} {Model} for {Satellite}
  {Imagery}.
\newblock In {\em The {Twelfth} {International} {Conference} on {Learning}
  {Representations}}, 2024.
\newblock arXiv:2312.03606.

\bibitem{li_generating_2020}
Yin Li and Da~Huang.
\newblock Generating {Hyperspectral} {Data} {Based} on {3D} {CNN} and
  {Improved} {Wasserstein} {Generative} {Adversarial} {Network} {Using}
  {Homemade} {High}-resolution {Datasets}.
\newblock August 2020.

\bibitem{lin_towards_2022}
Jue Lin, Gaurav Sharma, and Thrasyvoulos~N. Pappas.
\newblock Towards {Universal} {Texture} {Synthesis} by {Combining} {Texton}
  {Broadcasting} with {Noise} {Injection} in {StyleGAN}-2, March 2022.
\newblock arXiv:2203.04221 [cs].

\bibitem{liu_limited_1989}
Dong~C. Liu and Jorge Nocedal.
\newblock On the limited memory {BFGS} method for large scale optimization.
\newblock {\em Mathematical Programming}, 45(1):503--528, August 1989.

\bibitem{liu_diverse_2023}
Liqin Liu, Bowen Chen, Hao Chen, Zhengxia Zou, and Zhenwei Shi.
\newblock Diverse {Hyperspectral} {Remote} {Sensing} {Image} {Synthesis} {With}
  {Diffusion} {Models}.
\newblock {\em IEEE Transactions on Geoscience and Remote Sensing}, 61:1--16,
  2023.

\bibitem{mercier_characterization_2002}
G.~Mercier and M.~Lennon.
\newblock On the characterization of hyperspectral texture.
\newblock volume~5, pages 2584--2586 vol.5, February 2002.

\bibitem{mohandoss_generating_2020}
Tharun Mohandoss, Aditya Kulkarni, Daniel Northrup, Ernest Mwebaze, and Hamed
  Alemohammad.
\newblock Generating {Synthetic} {Multispectral} {Satellite} {Imagery} from
  {Sentinel}-2, December 2020.
\newblock arXiv:2012.03108 [cs, eess].

\bibitem{moradi_false-alarm_2018}
Saed Moradi, Payman Moallem, and Mohamad~Farzan Sabahi.
\newblock A false-alarm aware methodology to develop robust and efficient
  multi-scale infrared small target detection algorithm.
\newblock {\em Infrared Physics \& Technology}, 89:387--397, March 2018.

\bibitem{porebski_Comparison_2022}
Alice Porebski, Mohamed Alimoussa, and Nicolas Vandenbroucke.
\newblock Comparison of color imaging vs. hyperspectral imaging for texture
  classification.
\newblock {\em Pattern Recognition Letters}, 161:115--121, September 2022.

\bibitem{portilla_parametric_2000}
Javier Portilla and Eero~P. Simoncelli.
\newblock A {Parametric} {Texture} {Model} {Based} on {Joint} {Statistics} of
  {Complex} {Wavelet} {Coefficients}.
\newblock {\em International Journal of Computer Vision}, 40(1):49--70, October
  2000.

\bibitem{rabin_wasserstein_2012}
Julien Rabin, Gabriel Peyré, Julie Delon, and Marc Bernot.
\newblock Wasserstein {Barycenter} and {Its} {Application} to {Texture}
  {Mixing}.
\newblock In Alfred~M. Bruckstein, Bart~M. ter Haar~Romeny, Alexander~M.
  Bronstein, and Michael~M. Bronstein, editors, {\em Scale {Space} and
  {Variational} {Methods} in {Computer} {Vision}}, pages 435--446, Berlin,
  Heidelberg, 2012. Springer.

\bibitem{reed_adaptive_1990}
I.S. Reed and X.~Yu.
\newblock Adaptive multiple-band {CFAR} detection of an optical pattern with
  unknown spectral distribution.
\newblock {\em IEEE Transactions on Acoustics, Speech, and Signal Processing},
  38(10):1760--1770, October 1990.

\bibitem{rellier_texture_2004}
G.~Rellier, X.~Descombes, F.~Falzon, and J.~Zerubia.
\newblock Texture feature analysis using a gauss-{Markov} model in
  hyperspectral image classification.
\newblock {\em IEEE Transactions on Geoscience and Remote Sensing},
  42(7):1543--1551, July 2004.

\bibitem{rellier_gauss-markov_2002}
Guillaume Rellier, Xavier Descombes, Josiane Zerubia, and Frédéric Falzon.
\newblock A {Gauss}-{Markov} {Model} for {Hyperspectral} {Texture} {Analysis}
  of {Urban} {Areas}.
\newblock In {\em Proceedings of the 16 th {International} {Conference} on
  {Pattern} {Recognition} ({ICPR}'02) {Volume} 1}, {ICPR} '02, page 10692, USA,
  2002.

\bibitem{rombach_high-resolution_nodate}
Robin Rombach, Andreas Blattmann, Dominik Lorenz, Patrick Esser, and Björn
  Ommer.
\newblock High-{Resolution} {Image} {Synthesis} with {Latent} {Diffusion}
  {Models}.
\newblock In {\em 2022 {IEEE}/{CVF} {Conference} on {Computer} {Vision} and
  {Pattern} {Recognition} ({CVPR})}, pages 10674--10685. IEEE Computer Society.

\bibitem{sarkar_hyperspectral_2010}
Subhadip Sarkar and Glenn Healey.
\newblock Hyperspectral {Texture} {Synthesis} {Using} {Histogram} and {Power}
  {Spectral} {Density} {Matching}.
\newblock {\em IEEE Transactions on Geoscience and Remote Sensing},
  48(5):2261--2270, May 2010.
\newblock Conference Name: IEEE Transactions on Geoscience and Remote Sensing.

\bibitem{schweizer_hyperspectral_2000}
S.M. Schweizer and J.M.F. Moura.
\newblock Hyperspectral imagery: clutter adaptation in anomaly detection.
\newblock {\em IEEE Transactions on Information Theory}, 46(5):1855--1871,
  August 2000.

\bibitem{schweizer_efficient_2001}
S.M. Schweizer and J.M.F. Moura.
\newblock Efficient detection in hyperspectral imagery.
\newblock {\em IEEE Transactions on Image Processing}, 10(584-597), 2001.

\bibitem{szczap_flexible_2014}
F.~Szczap, Y.~Gour, T.~Fauchez, C.~Cornet, T.~Faure, O.~Jourdan, G.~Penide, and
  P.~Dubuisson.
\newblock A flexible three-dimensional stratocumulus, cumulus and cirrus cloud
  generator ({3DCLOUD}) based on drastically simplified atmospheric equations
  and the {Fourier} transform framework.
\newblock {\em Geoscientific Model Development}, 7(4):1779--1801, August 2014.
\newblock Publisher: Copernicus GmbH.

\bibitem{ulyanov_texture_2016}
Dmitry Ulyanov, Vadim Lebedev, Andrea Vedaldi, and Victor Lempitsky.
\newblock Texture networks: feed-forward synthesis of textures and stylized
  images.
\newblock In {\em Proceedings of the 33rd {International} {Conference} on
  {International} {Conference} on {Machine} {Learning} - {Volume} 48},
  {ICML}'16, pages 1349--1357, New York, NY, USA, 2016. JMLR.org.

\bibitem{ustyuzhaninov_texture_2016}
Ivan Ustyuzhaninov, Wieland Brendel, Leon~A. Gatys, and Matthias Bethge.
\newblock Texture {Synthesis} {Using} {Shallow} {Convolutional} {Networks} with
  {Random} {Filters}, May 2016.
\newblock arXiv:1606.00021 [cs].

\bibitem{wang_feature_2023}
Yi~Wang, Hugo~Hernández Hernández, Conrad~M. Albrecht, and Xiao~Xiang Zhu.
\newblock Feature {Guided} {Masked} {Autoencoder} for {Self}-supervised
  {Learning} in {Remote} {Sensing}, October 2023.
\newblock arXiv:2310.18653.

\bibitem{xiong_neural_2024}
Zhitong Xiong, Yi~Wang, Fahong Zhang, Adam~J. Stewart, Joëlle Hanna, Damian
  Borth, Ioannis Papoutsis, Bertrand~Le Saux, Gustau Camps-Valls, and
  Xiao~Xiang Zhu.
\newblock Neural {Plasticity}-{Inspired} {Foundation} {Model} for {Observing}
  the {Earth} {Crossing} {Modalities}, March 2024.
\newblock arXiv:2403.15356.

\bibitem{yu_unmixdiff_2024}
Yang Yu, Erting Pan, Yong Ma, Xiaoguang Mei, Qihai Chen, and Jiayi Ma.
\newblock {UnmixDiff}: {Unmixing}-{Based} {Diffusion} {Model} for
  {Hyperspectral} {Image} {Synthesis}.
\newblock {\em IEEE Transactions on Geoscience and Remote Sensing}, 62:1--18,
  2024.
\newblock Conference Name: IEEE Transactions on Geoscience and Remote Sensing.

\bibitem{zhang_unreasonable_2018}
Richard Zhang, Phillip Isola, Alexei~A. Efros, Eli Shechtman, and Oliver Wang.
\newblock The {Unreasonable} {Effectiveness} of {Deep} {Features} as a
  {Perceptual} {Metric}.
\newblock pages 586--595, 2018.

\end{thebibliography}
